\documentclass[acmsmall,screen,nonacm]{acmart}

\setcopyright{none}
\makeatletter
\@ifundefined{hyxmp@parse@acmart}{}{\let\hyxmp@parse@acmart\relax}
\makeatother

\usepackage{amsmath}
\usepackage{mathtools}
\usepackage{booktabs}
\usepackage{tabularx}
\usepackage{array}
\usepackage{multirow}
\usepackage{enumitem}
\usepackage{xcolor}
\usepackage{tikz}
\usepackage{microtype}
\usepackage{url}
\usepackage{placeins}
\usetikzlibrary{arrows.meta,positioning,fit,calc,shapes.geometric}

\hypersetup{
  keeppdfinfo=true,
  hidelinks,
  pdftitle={Language-Conditioned Token and Reasoning Efficiency in Large Language Models: A Paired Cross-Lingual Study Protocol},
  pdfauthor={Genliang Zhu; Chu Wang},
  pdfsubject={Protocol for paired multilingual evaluation under assigned observable trace-token budgets},
  pdfcreator={LaTeX}
}

\setlist[itemize]{leftmargin=1.25em}
\setlist[enumerate]{leftmargin=1.45em}
\newcolumntype{L}[1]{>{\raggedright\arraybackslash}p{#1}}
\newcolumntype{Y}{>{\raggedright\arraybackslash}X}

\definecolor{studyBlue}{HTML}{1D4ED8}
\definecolor{studyGreen}{HTML}{15803D}
\definecolor{studyAmber}{HTML}{B45309}
\definecolor{studyPurple}{HTML}{6D28D9}
\definecolor{studySlate}{HTML}{475569}

\newcommand{\Lin}{\ensuremath{L_{\mathrm{in}}}}
\newcommand{\Ltrace}{\ensuremath{L_{\mathrm{trace}}}}
\newcommand{\Lout}{\ensuremath{L_{\mathrm{out}}}}

\newcommand{\ITT}{\textsc{itt}}
\newcommand{\secref}[1]{Section~\ref{#1}}

\newcommand{\StudyStage}{design specified and preflight infrastructure verified}
\newcommand{\StudyAbstractStatusSentence}{No confirmatory model outcome is reported;
result fields remain disabled until the frozen evidence ledger passes
independent verification.}
\newcommand{\StudyConclusionStatusSentence}{The frozen design and verification
interface provide an auditable basis for the confirmatory study; empirical
estimates will be released after independent verification of the frozen
evidence ledger.}
 
\title[Language-Conditioned Token and Reasoning Efficiency]
{Language-Conditioned Token and Reasoning Efficiency in Large Language
Models: A Paired Cross-Lingual Study Protocol}

\author{Genliang Zhu}
\affiliation{%
  \institution{Accentrust}
  \city{Vancouver}
  \country{Canada}
}
\affiliation{%
  \institution{Georgia Institute of Technology}
  \city{Atlanta}
  \state{Georgia}
  \country{USA}
}
\email{research@accentrust.com}

\author{Chu Wang}
\affiliation{%
  \institution{Accentrust}
  \city{Vancouver}
  \country{Canada}
}
\affiliation{%
  \institution{University of Illinois Urbana-Champaign}
  \city{Urbana}
  \state{Illinois}
  \country{USA}
}

\begin{abstract}
Large language models incur language-dependent representation and inference
costs, but existing comparisons often conflate input language, assigned
observable-trace language, and answer realization. We specify a prospective
paired study that separates these interfaces while holding the semantic item,
checkpoint, and answer oracle fixed. The initial design instantiates 240 exactly
scored items rendered from templates in English and seven non-English languages,
three distinct-lineage open-weight checkpoints, three trace-token budgets, 22
input- and trace-language conditions, a fixed answer reserve, and a separately counted delimiter:
47,520 initial core runs before prospective sample-size selection. RQ1--RQ3
estimate input and trace effects by intention-to-treat with failure-inclusive
terminal accounting and test answer realization by cloning a sealed prefix and
runtime-native KV state into eight crossed branches. Pre-freeze independent
language review, fixed-form ASCII selectors, and code/surface/solver agreement
constrain the realization test. H1--H5 share one Holm family and a global
simultaneous component band. A secondary randomized experiment compares
one-long-attempt and complete \(K\)-short-attempt policies at equal trace allowance
under frozen seeds and oracle-blind aggregation; it is a full-policy contrast
because answer capacity differs. Outcomes include exact token spans,
correctness, latency, runtime-exposed memory, and qualified same-host
operating-system-reported energy over prespecified hardware rails. The protocol separates tokenizer
expansion, observable-trace cost, and answer-realization cost without treating
visible traces as internal cognition or operating-system estimates as physical
cross-device energy.
\StudyAbstractStatusSentence
\end{abstract}


\begin{document}
\maketitle
\hypersetup{
  keeppdfinfo=true,
  pdftitle={Language-Conditioned Token and Reasoning Efficiency in Large Language Models: A Paired Cross-Lingual Study Protocol},
  pdfauthor={Genliang Zhu; Chu Wang},
  pdfsubject={Protocol for paired multilingual evaluation under assigned observable trace-token budgets},
  pdfcreator={LaTeX}
}

\section{Introduction}
\label{sec:introduction}

Inference in a large language model is mediated by language at several
different boundaries. A user expresses a task in one language; a tokenizer
maps that expression into a model-specific sequence; a reasoning model may
generate an extended trace in the same or another language; and the model
finally realizes an answer for either a person or a machine. These boundaries
are economically and computationally consequential. Semantically parallel
texts can incur different API charges because their token counts differ
~\cite{ahia2023languages}. The same representation-length disparity can also
reduce usable context and increase processing latency
~\cite{petrov2023tokenizers}. Controlled tokenizer studies further associate
vocabulary coverage and tokenizer choice with downstream performance
~\cite{rust2021tokenizer,ali2024tokenizer}; compression alone is therefore not
a sufficient model of task quality~\cite{schmidt2024tokenization}.

Reasoning models add a second source of variation. Chain-of-thought prompting
and test-time computation can improve performance
~\cite{wei2022chain,kojima2022zeroshot}, yet multilingual gains are not uniform.
Cross-lingual prompting and multilingual instruction tuning can alter
reasoning accuracy and consistency~\cite{qin2023crosslingual,lai2024mcot}.
Explicit reasoning-language policies can additionally change generated-token
cost while preserving or changing accuracy~\cite{ahuja2025efficientxlang}.
Fixed-compute evaluation further shows that a test-time scaling policy need
not transfer uniformly across languages~\cite{son2025linguistic}. Recent
multilingual benchmarks observe cross-language differences in accuracy,
generated-trace length, and linguistic consistency~\cite{wang2025polymath}.
Controlled multilingual chain-of-thought evaluation separately shows that
input language and instructed reasoning language affect performance,
consistency, and faithfulness~\cite{zhao2026multilingualcot}.

These results establish that language matters, but they do not yield a single
causal explanation for an observed cost gap. Four quantities are commonly
entangled:

\begin{enumerate}
  \item the number of tokens required to represent an input;
  \item the length and content of the generated reasoning trace;
  \item the cost of realizing the final answer; and
  \item the model's success or failure under a fixed inference budget.
\end{enumerate}

A shorter response can reflect a better tokenizer, a premature answer, a
different reasoning strategy, noncompliance with the assigned language, or a
truncated run. Conversely, a larger token count need not imply proportionally
larger elapsed time because serving frameworks create distinct deployment
bottlenecks~\cite{fernandez2023framework}. Energy likewise depends on workload,
model, hardware, and inference optimization rather than token count alone
~\cite{fernandez2025energy,poddar2025sustainable}. Accordingly, a ranking based
only on mean tokens or on the ratio \(\textit{accuracy}/\textit{tokens}\)
cannot identify which boundary produced the difference or whether it remains
under a real compute budget.

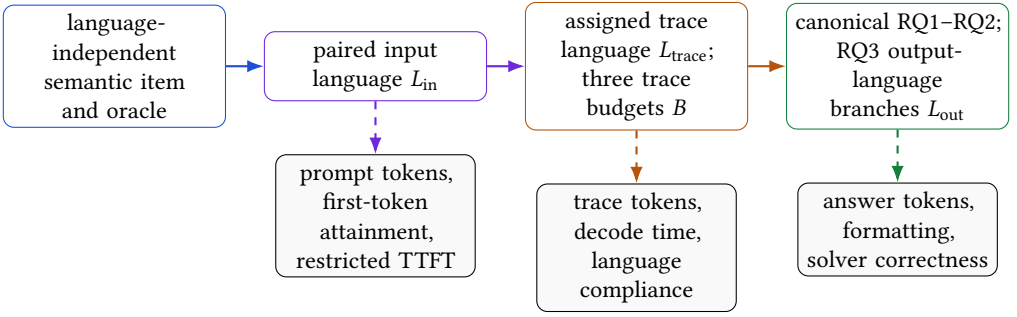
\begin{figure*}[t]
  \centering
  \begin{tikzpicture}[
    node distance=7mm and 5mm,
    box/.style={draw, rounded corners, minimum height=9mm, text width=27mm,
      align=center, font=\small, fill=white},
    metric/.style={draw, rounded corners, minimum height=8mm, text width=24mm,
      align=center, font=\small, fill=gray!5},
    arrow/.style={-{Latex[length=2.2mm]}, thick},
  ]
    \node[box, draw=studyBlue] (latent) {language-independent\\semantic item and oracle};
    \node[box, draw=studyPurple, right=of latent] (input) {paired input\\language $\Lin$};
    \node[box, draw=studyAmber, right=of input] (trace) {assigned trace language $\Ltrace$;\\three trace budgets $B$};
    \node[box, draw=studyGreen, right=of trace] (output) {canonical RQ1--RQ2;\\RQ3 output-language branches $\Lout$};
    \draw[arrow, studyBlue] (latent) -- (input);
    \draw[arrow, studyPurple] (input) -- (trace);
    \draw[arrow, studyAmber] (trace) -- (output);
    \node[metric, below=of input] (repr) {prompt tokens, first-token\\attainment, restricted TTFT};
    \node[metric, below=of trace] (reason) {trace tokens, decode time,\\language compliance};
    \node[metric, below=of output] (answer) {answer tokens, formatting,\\solver correctness};
    \draw[arrow, dashed, studyPurple] (input) -- (repr);
    \draw[arrow, dashed, studyAmber] (trace) -- (reason);
    \draw[arrow, dashed, studyGreen] (output) -- (answer);
  \end{tikzpicture}
  \caption{The planned design separates task presentation, assigned observable
  trace language, and answer realization. Input- and trace-language conditions form a complete
  paired schedule at all three assigned trace budgets per item--checkpoint
  block; execution order is randomized. Stage B fully crosses all eight answer
  languages after an exact parent-prefix/KV clone and randomizes only branch
  order.}
  \Description{A left-to-right pipeline begins with a language-independent
  semantic item, followed by paired input-language and assigned-trace-language
  conditions under each of three assigned observable trace-token budgets. The
  RQ1--RQ2 answer is canonical; primary Stage B clones one exact parent prefix
  and KV state before applying eight fixed-form branch-local language selectors
  in randomized execution order.
  Below the stages are token, timing,
  OS-reported energy, compliance, and correctness measures.}
  \label{fig:study-decomposition}
\end{figure*}

This paper specifies a paired protocol that will treat input language, assigned
generated-trace language, and output language as separate interfaces. The same
language-independent semantic item will be rendered in every registered input
language, evaluated by the same checkpoint, and scored by a deterministic
oracle. Primary input- and trace-language conditions will form a complete paired schedule.
Every item--checkpoint block will receive exactly three assigned observable
trace-token budgets, \(b_L,b_M,b_H\), and the complete schedule of 22
input- and trace-language conditions within every budget. A separate fixed
answer-token reserve \(a_0\) is
shared across conditions, while the structural trace--answer delimiter is
counted separately. RQ1--RQ2 use a language-neutral answer. Primary RQ3 will
clone the same sealed pre-treatment serialized prefix and runtime-native KV
state into eight fully crossed branches, randomize branch order, and append a
fixed-form ASCII selector whose meaning is defined in a shared codebook that
must pass qualified two-reviewer validation before design freeze and is then
sealed into the parent; textual replay in fresh requests
is robustness-only. Randomized secondary RQ5 will compare a stochastic
one-long-attempt policy with a complete \(K\)-short-attempt policy at equal total
trace allowance under frozen seeds and oracle-blind aggregation. Because each
short attempt has its own answer reserve, RQ5 estimates the complete policies
rather than a pure trace-allocation effect. Core and RQ3 use a qualified
deterministic greedy decoding profile. The protocol will record token counts,
time to first token (TTFT), decode time, wall time, and runtime-exposed peak
memory when available without profiler
distortion, and operating-system-reported energy over prespecified hardware
rails on one fixed host. Time and energy will be outcomes under assigned trace budgets,
not post-treatment matching variables.
Figure~\ref{fig:study-decomposition} summarizes the separation.

The protocol makes four methodological contributions. Empirical conclusions
are reported only after the registered evidence gates are satisfied.

\begin{itemize}
  \item \textbf{A paired finite-set estimand.} It will separate \(\Lin\) and
  assigned \(\Ltrace\) within the same checkpoint, semantic item, and exogenous
  trace-budget operating point; \(\Lout\) will be evaluated in a separately
  frozen primary continuation experiment from identical pre-treatment states.
  \item \textbf{A cost decomposition beyond tokens.} It will measure prompt,
  trace, and answer token spans jointly with latency, same-host OS-reported
  energy estimates over prespecified hardware rails and, where exposed without profiler distortion,
  memory, following systematic reporting
  principles for machine-learning energy studies~\cite{henderson2020energy}
  and request-aligned measurement practice for LLM inference
  ~\cite{niu2026tokenpowerbench}.
  \item \textbf{An analysis robust to instruction noncompliance.} Assigned
  trace language will be analyzed by intention-to-treat. Detected language
  mixing will be reported as an outcome because reasoning models commonly mix
  languages~\cite{wang2025languagemixing}, and bilingual experiments show that
  the effect of such mixing is condition-dependent rather than uniformly
  harmful~\cite{li2025bilingualmixing}.
  \item \textbf{A reproducible multilingual task layer.} Typed latent items,
  executable oracles, native-language template review, timestamped immutable
  manifests, and an independent checker will provide a contamination-resistant
  complement to public benchmarks.
\end{itemize}

The paper will not treat English as an intrinsically superior reasoning
language, equate visible traces with internal cognition, or generalize from the
registered languages to all language communities. Checkpoint and language
comparisons will describe the finite registered set; confirmatory contrasts
will be indexed by assigned observable trace-token budget. This boundary is
essential because generated explanations can be unfaithful
~\cite{turpin2023unfaithful} and recent evidence does not support a simple
equivalence between surface language and latent reasoning
~\cite{liu2026latent}.
\section{Related Work}
\label{sec:related-work}

\subsection{Tokenization, representation cost, and fairness}

Subword tokenization enables open-vocabulary modeling through reusable units;
byte-pair encoding and SentencePiece are foundational examples
~\cite{sennrich2016bpe,kudo2018sentencepiece}.  A multilingual vocabulary,
however, allocates finite capacity across scripts, morphologies, and language
frequencies.  Controlled studies associate tokenizer coverage and fertility
with monolingual downstream performance~\cite{rust2021tokenizer}, while
vocabulary allocation and cross-language overlap can help some tasks and harm
others~\cite{limisiewicz2023tokenization}.  Large-scale tokenizer ablations
show that vocabulary choice affects training cost and quality, but also caution
that fertility and parity alone are unreliable proxies for downstream
performance~\cite{ali2024tokenizer}.

The deployment consequences are independently established.  Commercial APIs
can charge different amounts for semantically parallel content because token
counts differ by language~\cite{ahia2023languages}; tokenizer-induced length
differences also affect latency and effective context capacity
~\cite{petrov2023tokenizers}.  Recent work improves cross-language parity at
tokenizer construction time~\cite{foroutan2026parity} or changes segmentation
across conventional token boundaries~\cite{tanase2026supratok}.  These studies
motivate our representation-cost outcomes.  They also motivate a strict
separation between tokenization and reasoning: minimizing the number of tokens
does not necessarily maximize task quality~\cite{schmidt2024tokenization}.

\subsection{Multilingual generated reasoning}

Chain-of-thought and zero-shot reasoning prompts established that additional
generated computation can improve performance on multi-step tasks
~\cite{wei2022chain,kojima2022zeroshot}.  Multilingual extensions use
cross-lingual prompts or instruction tuning to transfer reasoning behavior
~\cite{shi2023multilingualcot,qin2023crosslingual,lai2024mcot}.  More recent work directly chooses or
regularizes reasoning language.  EfficientXLang evaluates whether
cross-lingual reasoning can reduce trace length while retaining mathematical
accuracy~\cite{ahuja2025efficientxlang}; unified-logic approaches similarly
optimize cross-language trace length and latency~\cite{zhang2026lesslanguages}.
English can function as a useful pivot, but translation and language mismatch
can also introduce errors~\cite{saji2026linguafranca}.
Multilingual negotiation experiments further show that reasoning quality and
incurred cost can vary jointly with language and interaction structure
~\cite{hakimov2026price}, reinforcing the need for paired quality--cost
estimands rather than an output-length leaderboard.

Two findings are especially important for identification.  First, input language and
the language assigned for an observable reasoning trace can both affect accuracy, consistency, language
compliance, and faithfulness~\cite{zhao2026multilingualcot}.  Second, reasoning
models frequently mix languages~\cite{wang2025languagemixing}.  Bilingual
experiments separately show that forcing one surface language can reduce
performance in some conditions~\cite{li2025bilingualmixing}.  We therefore randomize
the assignment but analyze its total policy effect by \ITT.  Actual detected
language is a post-treatment variable and cannot be used as if it were
randomized.  Complementary layer-level interventions suggest that multilingual
models can use different internal language-specific pathways from input through
generation~\cite{zhao2024multilingualworkflow}; those mechanism results do not,
however, identify the effect of an externally assigned trace-language policy on
tokens, latency, energy, or correctness.
Representation-engineering work further separates cross-lingual reasoning
enhancement from target-language output anchoring and evaluates input--output
language consistency~\cite{li2026mrre}.  This supports treating output control
as a distinct intervention surface; it does not, by itself, identify the
output-language branch contrast from an identical pre-treatment state studied here.

\subsection{Test-time scaling and multilingual benchmarks}

Inference-time budgets introduce a quality--cost curve rather than a single
operating point. Compute-optimal inference work shows more generally that the
value of extra test-time computation depends on item difficulty and allocation
strategy, so a single unrestricted operating point is not a sufficient
efficiency characterization~\cite{snell2025testtime}. Multilingual mathematical
evaluation shows that test-time
scaling gains can vary across languages even when inference FLOPs are held
fixed~\cite{son2025linguistic}.  PolyMath reports variation in accuracy,
generated-trace length, and input--output language consistency across eighteen
languages~\cite{wang2025polymath}.  MMLU-ProX contributes professionally
reviewed parallel questions across languages and emphasizes expert validation
and terminology consistency~\cite{xuan2025mmluprox}.  Macaron addresses a
different validity threat: translating English-centered situations does not
produce culturally or linguistically native reasoning data
~\cite{elsetohy2026macaron}.

Self-consistency established sample-and-aggregate decoding by drawing diverse
reasoning paths and selecting the most consistent final answer
~\cite{wang2023selfconsistency}.  RQ5 studies a different object: a randomized
contrast between complete one-long-attempt and \(K\)-short-attempt policies at equal
total trace allowance, with repeated request, answer, aggregation, latency,
and qualified energy costs retained.

Public benchmarks provide necessary external validity, but they are not ideal
for the primary estimand here.  Items may differ in translation quality,
cultural presupposition, contamination exposure, or answer realization.
Consequently, our primary task layer begins with a typed language-independent
representation and executable answer oracle.  Public multilingual benchmarks
form a separately reported external-validation stratum.

\subsection{Evaluation uncertainty and experimental design}

NLP experiments are often underpowered for the small effects they seek to
detect, increasing both false-negative risk and the chance that statistically
significant estimates exaggerate the underlying effect
~\cite{card2020power}. Multilingual and multitask evaluation introduces an
additional external-validation problem: both data sampling and model-side variation
can materially affect rankings and pairwise differences
~\cite{saleva2025uncertainty}. These findings motivate prospective simulation
of the complete registered design rather than a post hoc power calculation.

The paired schedule also has crossed dependence. Multiple language conditions
share a semantic item, the English reference C4 is reused across contrasts,
and latency and energy observations share thermal execution blocks. Work on
dependent online experiments shows that ignoring more than one clustering unit
can make uncertainty intervals too narrow and that multiway resampling may be
needed~\cite{bakshy2013dependent}. We therefore preserve semantic-item and
thermal-block dependence, as well as shared-reference covariance, in the
registered resampling procedure; the cited online-experiment result motivates
the dependence audit rather than validating this study's estimator by itself.
Resampling-based multiple-testing procedures can use the joint dependence of
studentized statistics to construct family-wise error-controlling critical
values~\cite{romano2005stepdown}.  This motivates the common
max-\(\lvert t\rvert\) layer; the study-specific crossed resampling algorithm,
finite-set weights, and shared-reference handling are separately frozen and
independently verified.

\subsection{Inference latency and energy}

Tokens are workload units, not universal compute units.  Framework overhead
can cause nominally efficient architectures to perform differently in deployed
latency~\cite{fernandez2023framework}. Production-oriented serving work further
shows that prefill and autoregressive decode have distinct compute and latency
profiles, and that batching and scheduling alter their observed trade-off
~\cite{agrawal2024sarathi}. LLM inference energy varies across
models, hardware, inference engines, and workloads
~\cite{poddar2025sustainable}.  Input/output length, batching, and efficiency
optimizations further change measured inference energy
~\cite{fernandez2025energy}.  Systematic energy
reporting therefore requires explicit measurement boundaries and hardware
metadata~\cite{henderson2020energy}.  TokenPowerBench further separates
prefill and decode phases and aligns power samples with individual requests
~\cite{niu2026tokenpowerbench}.

We apply those principles to a multilingual intervention.  The study does not
infer joules from token counts or compare incompatible device-level and
board-level power estimates.  It will report accuracy and failure-inclusive
unconditional costs at three exogenously assigned observable trace-token
budgets, with an independent fixed answer reserve.  For each cost it will also
report the registered joint event of correctness and remaining below a cost
threshold.  Views that condition on realized time or energy remain descriptive
rather than being presented as fixed-cost causal effects.

\subsection{Difference from the nearest work}

The closest efficiency study establishes that a cross-lingual reasoning policy
can change generated-token cost and mathematical accuracy
~\cite{ahuja2025efficientxlang}. A multilingual negotiation study likewise
measures performance--cost trade-offs and reports model-dependent switching to
English in observable reasoning traces~\cite{hakimov2026price}; its domain,
unpaired interface assignment, and cost boundary differ from ours. A separate
controlled evaluation shows that
input and instructed reasoning language affect multilingual performance,
consistency, and faithfulness~\cite{zhao2026multilingualcot}; multilingual
test-time scaling is also heterogeneous~\cite{son2025linguistic}.  Our
contribution is a paired, interface-level efficiency study rather than a
language-ranking benchmark.  It combines: (i) paired separation of
input, assigned trace, and output language; (ii) a language-independent,
exactly scored primary task layer; (iii) request latency and qualified same-host
operating-system-reported energy measurement over prespecified hardware rails; (iv) assigned trace-budget
accuracy operating points with descriptive realized-cost views; and (v) ITT treatment of
language noncompliance. Together these controls are designed to attribute a
measured cost difference to a defined interface more narrowly than an
uncontrolled mixture of representation, reasoning, and realization. Primary
answer realization is
identified by branching from an identical pre-treatment serialized prefix and
runtime-native KV state, then append-prefilling a randomized, fixed-form
language-control suffix whose codebook must pass the pre-freeze language-review
gate and whose token and prefill costs are measured;
fresh-request textual replay is retained only as a robustness analysis.
\section{Measurement Model and Estimands}
\label{sec:measurement-model}

\subsection{Units, schedule, and treatment assignment}

Let \(i\) denote a language-independent semantic item, \(m\) one frozen
checkpoint in the registered finite set, \(b\in\{b_L,b_M,b_H\}\) one of three
assigned observable trace-token budgets, and \(c\) an input- and trace-language
experimental condition. A
separate fixed answer-token reserve \(a_0\) is identical across conditions; the
structural trace--answer delimiter belongs to neither allowance and is counted
separately. Neither allowance is a wall-time or energy cap. The primary
treatment record is

\begin{equation}
  Z=(c,b), \qquad c=(\Lin,\Ltrace,\Lout=\mathrm{canonical}),
  \qquad a=a_0.
  \label{eq:treatment}
\end{equation}

One stateless invocation for a fixed item, checkpoint, cell, trace budget, and
decoding seed is the atomic scheduled identity. Every item--checkpoint pair
receives all three trace budgets, and every item--checkpoint--budget block
receives the complete input- and trace-language schedule. A committed permutation assigns
each \((i,m,c,b,\mathrm{seed})\) identity to one thermal execution slot, balanced
within checkpoint, task family, and registered difficulty. It never removes a
budget or language cell. Semantic item is the content-dependence cluster;
thermal execution block is an additional latency/energy cluster. Randomized run
order controls nuisance variation and is not itself a scientific treatment.
Core RQ1--RQ2 uses one frozen greedy profile with sampling disabled,
temperature zero, and top-\(p\) one. Its seed binds serialization and execution
provenance but is not an inferential factor. Before freeze, each checkpoint must
reproduce byte-identical token output on a registered repeat suite under this
profile.

For each of seven registered non-English languages \(\ell\), the primary
schedule contains C1--C3. C4 is generated once per
item--checkpoint--budget--seed and shared by all seven comparisons.

\begin{table}[t]
  \centering
  \caption{Primary schedule. C1--C3 are run for each non-English language
  \(\ell\); C4 is one shared baseline.}
  \label{tab:primary-cells}
  \begin{tabular}{lllll}
    \toprule
    Cell & Scope & \(\Lin\) & assigned \(\Ltrace\) & \(\Lout\) \\
    \midrule
    C1 & per \(\ell\) & \(\ell\) & \(\ell\) & canonical \\
    C2 & per \(\ell\) & \(\ell\) & English & canonical \\
    C3 & per \(\ell\) & English & \(\ell\) & canonical \\
    C4 & shared & English & English & canonical \\
    \bottomrule
  \end{tabular}
\end{table}

At the initial planning value of 240 items, with three checkpoints, three trace
budgets, and one decoding seed, the RQ1--RQ2 schedule contains

\begin{equation}
  240\times3\times3\times(7\times3+1)=47{,}520
  \label{eq:primary-run-count}
\end{equation}

runs. Primary RQ3 continuations, repeated-seed robustness, and randomized RQ5
attempt allocation are counted separately. Trace budgets and language cells are
complete controlled repeated conditions; their execution order is randomized
for nuisance control.

\subsection{Finite registered-set estimands}

Let \(Y^{(k)}_{imc}(b)\) be outcome \(k\) for item \(i\), checkpoint \(m\), cell
\(c\), and assigned trace budget \(b\). At fixed checkpoint, language, and
trace budget,
input-language contrasts are C1--C3 and C2--C4; assigned-trace contrasts are
C1--C2 and C3--C4. Let \(N\) denote the final power-frozen RQ1--RQ2 item count;
the initial planning value is \(N=240\). For example,

\begin{align}
  \Delta^{A}_{\mathrm{in},m}(\ell;\ell,b)
    &= \frac{1}{N}\sum_i
       \left[A_{im,\mathrm{C1}(\ell)}(b)-
             A_{im,\mathrm{C3}(\ell)}(b)\right], \\
  \Delta^{A}_{\mathrm{trace},m}(\ell;\ell,b)
    &= \frac{1}{N}\sum_i
       \left[A_{im,\mathrm{C1}(\ell)}(b)-
             A_{im,\mathrm{C2}(\ell)}(b)\right].
  \label{eq:paired-effects}
\end{align}

Complete trace-budget crossing identifies operating-point contrasts within the
same item--checkpoint blocks; paired cells remove semantic-item variation.
Checkpoints and languages are not probability samples. Results will report
checkpoint- and language-specific estimates first. The registered aggregate
is an equal-weight arithmetic mean over the three checkpoints and seven
non-English languages, retaining C4 once and preserving its shared covariance.
It is a finite-set summary, not a model-, resource-, or world-language
population effect.

\subsection{Primary answer realization}

RQ3 is a separately executed primary Stage-B experiment on disjoint items. Its
finite target frame crosses all 22 input- and trace-language conditions, three trace budgets,
and three checkpoints for \(n_3\) power-selected items. Execution uses a
balanced-incomplete probability panel: for every item--checkpoint pair, an
outcome-blind restricted draw selects one of 21 non-English-language--budget
pairs and executes the corresponding C1, C2, C3, and C4 parents. Counts differ
by at most one within checkpoint--task-family--difficulty strata, with gaps
balanced across checkpoints. Specified C1--C3 parents have inclusion
probability \(1/21\); C4 at a specified budget has probability \(1/3\). The
draw and Horvitz--Thompson/Hájek weights freeze before generation.

The target frame contains \(198n_3\) parents; the executed panel contains
\(12n_3\) parents and, before eligibility losses, \(96n_3\) output branches.
An eligible Stage-A run
completes its observable trace, appends and separately counts the common
structural delimiter, then stops at the registered control-injection boundary
before any output-language control or answer token, and exposes the complete
serialized prefix-token sequence together with a runtime-native, immutable KV
snapshot. Before any branch runs, the runner seals the prefix digest, KV
digest, boundary, parent-state identity, checkpoint/runtime identity, eight
branch descriptors, and randomized branch order.

The sealed parent contains one shared codebook that must pass qualified
two-reviewer validation before design freeze and maps eight equal-status
fixed-width ASCII codes to the registered output languages. Each
Stage-B arm is restored from a byte-identical copy-on-write clone and then
receives the same fixed-form ASCII selector template with only the registered
language code changed. A single control mechanism is frozen across checkpoints.
Its exact token IDs and span are recorded, and no
arm may inherit sampler state, conversational history, or mutable cache changes
from another arm. Thus the branches share an exact pre-treatment parent state;
the protocol does not claim that their post-control KV states are identical.
All eight branches use the same deterministic greedy profile as the parent;
only their execution order is randomized.

RQ3 freezes a two-field controlled short-answer schema. A byte-stable ASCII
\texttt{answer\_code} carries the canonical solver payload, and
\texttt{answer\_text} must instantiate a localized template that has passed
the same pre-freeze review gate for the assigned output language and task
family. Semantic preservation requires a
valid schema, correct code, valid localized surface form, and agreement among
the surface-extracted slot, code, and independent solver. A correct code paired
with an invalid or contradictory localized sentence therefore fails
preservation. This estimand concerns controlled short-answer realization, not
open-ended multilingual generation.

For parent \(j\), selection indicator \(S_j\), mechanical pre-treatment
eligibility \(E_j\), inclusion probability \(\pi_j\), output language
\(\ell\), English reference branch \(e\), and endpoint transform \(g_k\), the
registered design-weighted contrast is
\begin{equation}
  \widehat{\theta}_{\ell k} =
  \frac{\sum_j S_jE_j\pi_j^{-1}
    \{g_k(Y_{j\ell})-g_k(Y_{je})\}}
       {\sum_j S_jE_j\pi_j^{-1}}.
  \label{eq:rq3-hajek}
\end{equation}
Binary components use risk differences. Cost ratios are ratios of the
corresponding weighted marginal means, not means of parent-level ratios.

H4 estimates conditional answer realization in the observed
pre-treatment-eligible states of that frozen target frame. It records selector tokens and
prefill time as descriptive design diagnostics. Its inferential components are
answer tokens and decode time, total branch wall time, qualified
incremental energy from control injection through the terminal event, format
validity, and solver-verified semantic preservation. Every branch receives the
same fixed \(a_0\) answer reserve and terminal precedence. Planned but ineligible
states remain in the attrition ledger and are not replaced. Eligibility,
sample size, weights, branch construction, clone validation, endpoints, and H4
freeze before Stage-A collection. A fresh-request adapter that replays the exact
serialized text is retained only as a labeled robustness analysis; it repeats
prefill and is never pooled with, or used to rescue, the result based on
identical pre-treatment states.

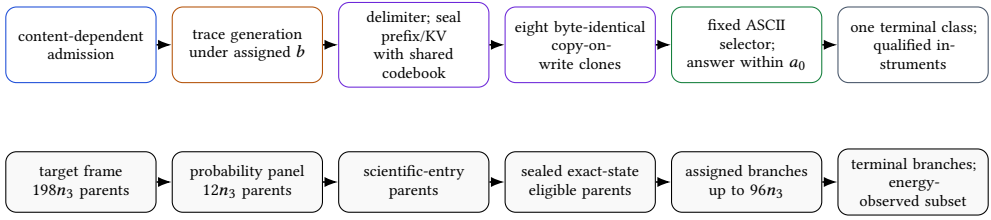
\begin{figure*}[t]
  \centering
  \begin{tikzpicture}[
    node distance=2mm,
    stage/.style={draw, rounded corners, minimum height=10mm, text width=17.5mm,
      align=center, font=\tiny, fill=white},
    denom/.style={draw, rounded corners, minimum height=8mm, text width=17.5mm,
      align=center, font=\tiny, fill=gray!5},
    flow/.style={-{Latex[length=1.8mm]}, semithick},
  ]
    \node[stage, draw=studyBlue] (entry) {content-dependent\\admission};
    \node[stage, draw=studyAmber, right=of entry] (trace) {trace generation\\under assigned $b$};
    \node[stage, draw=studyPurple, right=of trace] (seal) {delimiter; seal prefix/KV\\with shared codebook};
    \node[stage, draw=studyPurple, right=of seal] (clone) {eight byte-identical\\copy-on-write clones};
    \node[stage, draw=studyGreen, right=of clone] (branch) {fixed ASCII selector;\\answer within $a_0$};
    \node[stage, draw=studySlate, right=of branch] (terminal) {one terminal class;\\qualified instruments};
    \draw[flow] (entry) -- (trace);
    \draw[flow] (trace) -- (seal);
    \draw[flow] (seal) -- (clone);
    \draw[flow] (clone) -- (branch);
    \draw[flow] (branch) -- (terminal);

    \node[denom, below=9mm of entry] (frame) {target frame\\$198n_3$ parents};
    \node[denom, right=of frame] (sample) {probability panel\\$12n_3$ parents};
    \node[denom, right=of sample] (entered) {scientific-entry\\parents};
    \node[denom, right=of entered] (eligible) {sealed exact-state\\eligible parents};
    \node[denom, right=of eligible] (branches) {assigned branches\\up to $96n_3$};
    \node[denom, right=of branches] (observed) {terminal branches;\\energy-observed subset};
    \draw[flow] (frame) -- (sample);
    \draw[flow] (sample) -- (entered);
    \draw[flow] (entered) -- (eligible);
    \draw[flow] (eligible) -- (branches);
    \draw[flow] (branches) -- (observed);
  \end{tikzpicture}
  \caption{RQ3 state, intervention, and denominator boundaries. The shared
  codebook is sealed before branching; the eight clones execute in randomized
  order, and only the fixed selector and answer are newly processed per branch.
  Every denominator is reported, and ineligible
  sampled parents are retained in attrition rather than replaced.}
  \Description{The top row shows scientific admission, trace generation,
  delimiter and prefix/KV sealing with a shared codebook, eight copy-on-write
  clones, a fixed selector and answer, and terminal accounting. The bottom row
  shows target-frame parents, sampled parents, scientific-entry parents,
  exact-state-eligible parents, assigned branches, and terminal branches with a
  separately identified energy-observed subset.}
  \label{fig:rq3-boundaries}
\end{figure*}

\subsection{Intention-to-treat and terminal outcomes}

Detected trace language \(D\) occurs after assignment. Primary trace-policy
estimands therefore use assigned \(\Ltrace\), retaining mixing,
noncompliance, empty traces, truncation, content-dependent admission rejection,
timeout, and OOM. Only a solver-verified canonical answer is pass@1 success.
All other scientific terminal classes are failures. Trace-budget exhaustion is
recorded separately and is not itself a failure when the frozen handoff rule
produces a valid answer within \(a_0\).

The scientific outcome boundary begins when a treatment-bound serialized
request enters content-dependent runtime admission, not only after the first
model kernel runs. Every such request receives one terminal class under a
frozen precedence rule: context overflow, input rejection, content-policy
rejection, OOM, timeout, answer-reserve exhaustion without a valid answer,
completed malformed, completed incorrect, completed correct, runtime error,
runtime crash, or operator-cancelled deviation. The final three abnormal
infrastructure classes remain failure-inclusive and trigger the registered
stop/deviation review; they are never rerun for improvement. Trace-budget
exhaustion is a nonexclusive process event before forced handoff, not a
competing terminal class. Terminal class is modeled multinomially and reported
in absolute counts; compliance is separate.

Only a demonstrably treatment-independent transport, host, or bootstrap failure
before content-dependent admission has no scientific outcome and may receive
the single registered replacement. Content-dependent context/length/policy
rejection is never replaceable. Evidence-integrity failure is not relabeled as
a model failure: it places the affected claim set on protocol-deviation hold.

Trace and answer counts include partial failure tokens. Their registered
analysis is two-part: zero-versus-positive logistic plus a positive count
model. The H3 token component is their composed unconditional marginal mean,
\(\Pr(Y>0)\mathrm{E}[Y\mid Y>0]\), over all model outcomes; its primary effect
is the paired natural-unit difference. The two parts are mechanism summaries,
not extra H3 components. A count ratio is eligible only when both unconditional
marginal means are strictly positive. TTFT is paired with first-token attainment; restricted TTFT equals the
registered timeout when no token arrives. Wall time and total-request gross
energy include post-admission scientific failures through their observed terminal boundary.
No latent completion cost is imputed.

\subsection{Exact token-span accounting}

The runtime records complete serialized input and generated token-ID arrays.
Registered sentinels define disjoint index ranges for system text, task
rendering, policy instruction, format instruction, structural overhead,
observable trace, trace--answer delimiter, RQ3 branch-appended control suffix,
and answer. The RQ3 control suffix is an append-prefill input span after the
sealed parent sequence; it is not relabeled as generated output. Counts are
lengths of runtime token-array ranges, not sums from independently tokenized
fragments:

\begin{equation}
  T_j=T^{\mathrm{prompt}}_j+T^{\mathrm{generated}}_j
      =\sum_{h\in\mathcal{H}_{\mathrm{prompt}}}|I_{jh}|
       +\sum_{h\in\mathcal{H}_{\mathrm{generated}}}|I_{jh}|.
  \label{eq:token-decomposition}
\end{equation}

The index sets are disjoint and cover the runtime arrays exactly. Tokens per
UTF-8 byte and Unicode grapheme are descriptive normalizations. A checkpoint
or runtime without exact IDs and an observable, assignable trace--answer
boundary is ineligible for primary trace analysis.

For an RQ3 branch \(r\), the separately verified incremental decomposition is
\begin{equation}
  T^{\mathrm{branch}}_r
  =T^{\mathrm{control\mbox{-}append}}_r+T^{\mathrm{answer}}_r,
  \label{eq:rq3-token-decomposition}
\end{equation}
where the first term indexes the uniquely appended selector IDs and the second
the uniquely generated answer IDs. Together with the sealed parent digest,
these spans cover every token position appended or generated after the branch
point; they do not recount positions already represented by the restored KV
state.

\subsection{Time and operating-system-reported energy outcomes}

Primary model wall time runs from content-dependent admission to the terminal
event. Pre-admission transport/bootstrap overhead is recorded separately
as descriptive infrastructure evidence. Trace- and answer-decode timing require
validated boundaries. If its frozen qualification gate
passes, the energy outcome is the same-host total-request gross estimate over
the operating-system rails explicitly named in the frozen qualification receipt,

\begin{equation}
  E_j^{\mathrm{gross}}=\int_{t_{0j}}^{t_{1j}}
  P_{\mathrm{included\ rails}}(t)\,dt.
  \label{eq:energy}
\end{equation}

This operating-system-reported estimate over prespecified hardware rails is
neither total wall-plug energy,
a calibrated physical-meter reading, nor a cross-device quantity.
Prefill/decode phase estimates are exploratory
and appear only if the stricter phase-resolution gate passes. Idle-adjusted
net energy is secondary and may be negative, so it uses natural-scale
differences and an identity-link sensitivity model. A missing or corrupt power
stream is instrumentation missingness, never zero energy and never imputed
from tokens. Before outcomes, the qualification receipt freezes a conservative
finite \(E_{\max}\) from the validated platform-power ceiling and request
timeout. Missing gross energy is propagated through worst-case
\([0,E_{\max}]\) bounds for means/ratios and lower/upper joint-curve bounds. If
energy remains primary with allowed missingness, its H3/H4 contribution uses the
supremum omnibus \(p\)-value over all admissible completions and an independently
reproduced optimization certificate. A named energy or efficiency claim
requires every extreme-case simultaneous bound to qualify. If the bound,
coverage gate, or certificate fails, the registered fallback removes energy
from H3/H4 and efficiency claims while leaving one non-energy H3 omnibus and the
five-hypothesis Holm family unchanged.

Missing registered latency is handled symmetrically. The frozen timeout defines
a conservative finite \(T_{\max}\); admissible values lie in
\([0,T_{\max}]\). If missing latency is allowed by its coverage gate, H1, H3,
and H4 time components, their global-band intervals, and their omnibus
contribution use least-favorable bounds and the supremum \(p\)-value over
admissible completions, with an independent optimization certificate. Failure
of the bound, gate, or certificate activates the frozen fallback that removes
only the affected time components while retaining the H-level test and
five-test Holm family. An energy or latency fallback rules out an unqualified
efficiency claim; only qualifying endpoint-specific statements remain eligible.

\subsection{Assigned trace-budget operating points}

For language policy \(p\), the confirmatory accuracy curve is indexed by the
assigned observable trace-token budget,

\begin{equation}
  g_p(b)=\Pr\!\left(A=1\mid Z_{\mathrm{language}}=p,\ B=b\right).
  \label{eq:budget-curve}
\end{equation}

Here and below, \(\Pr\) denotes the registered empirical distribution induced
by uniform item weighting and the stated finite-set or design weights over the
frozen checkpoints, cells, and sampled RQ3 parents. It does not posit random
sampling of languages or checkpoints from a larger population.

Each operating point also reports unconditional observable trace tokens, total
generated-sequence tokens (trace plus delimiter plus answer), restricted TTFT,
wall time, qualified gross same-host OS-reported energy estimate, and terminal-state
probabilities. For every registered realized cost \(C\), the failure-inclusive
joint success--cost curve is

\begin{equation}
  G^{C}_{p,b}(c)=\Pr\!\left(A=1\ \land\ C\leq c
    \mid Z_{\mathrm{language}}=p,\ B=b\right).
  \label{eq:joint-success-cost}
\end{equation}

This quantity is never replaced by
\(\Pr(A=1\mid C\leq c,p,b)\), which conditions on a post-treatment realized
cost. Time and energy are measured outcomes, not fixed-time or fixed-energy
interventions. Pareto sets and realized-cost AUCs, if shown, remain descriptive
and receive no causal interpretation.
\section{Paired Task and Language Construction}
\label{sec:corpus}

\subsection{Typed latent tasks}

The primary dataset will be generated from typed, language-independent semantic
objects.  This design will keep the answer, difficulty controls, and logical
relations constant before any natural-language rendering occurs.  Four task
families provide complementary reasoning demands:

\begin{enumerate}
  \item \textbf{Arithmetic and algebra}: integer, proportional, and small
  symbolic systems with executable solutions;
  \item \textbf{Constraint and deductive reasoning}: ordering, selection, and
  satisfiability instances with unique or explicitly enumerated answers;
  \item \textbf{Planning and path selection}: weighted graphs, schedules, and
  resource-constrained plans evaluated by an independent solver; and
  \item \textbf{Synthetic multi-hop composition}: typed fact sets in which
  requested relations require two or more deterministic joins.
\end{enumerate}

The implemented candidate generator realizes two subtypes per family---state
transitions and linear/proportional systems; partial orders and finite-domain
constraint satisfaction; weighted graphs and resource scheduling; Horn-rule
inference and relational joins---at three structural difficulty levels with ten
items per subtype--level cell, for 240 latent items. An independently
implemented oracle currently verifies a unique answer for all 240 items; 1,380
of 1,380 registered constraint mutants are rejected, direct answer disclosure
is zero, and 20 independent seed/property runs verify 4,800 generated items and
27,600 mutants. These are generator and corpus-integrity tests, not model
outcomes. The final confirmatory instances are still derived only after the
external-beacon freeze described below.

Every latent item will have an immutable identifier, task-family label,
registered difficulty, typed representation, canonical answer, and content
digest.  The generator and the independent checker will use separately
implemented oracles.  An item will be admitted only when both agree and when mutation tests demonstrate that
changing a registered constraint can change or invalidate the answer.

The confirmatory set will be generated only after the generator, templates, and
analysis plan are frozen.  The canonical freeze manifest will identify a future
round of an external public randomness beacon.  Its not-yet-known value and a
precommitted domain separator will derive the seed after the manifest and
generator digest receive an external timestamp.  Pilot and confirmatory items
will use disjoint namespaces and cannot share an identifier or latent
representation.

\subsection{Language selection}

The confirmatory set contains exactly eight total languages: US English (EN;
\texttt{en-US}), Mainland Standard Written Chinese in simplified characters
(ZH; \texttt{zh-Hans-CN}), Spain Spanish (ES; \texttt{es-ES}), Modern Standard
Arabic (AR), India Hindi in Devanagari (HI; \texttt{hi-IN}), Japan Japanese
(JA; \texttt{ja-JP}), Tanzania Swahili (SW; \texttt{sw-TZ}), and Thailand Thai
(TH; \texttt{th-TH}). English is the common reference, leaving seven registered
non-English contrasts.  The codes denote study treatments, not homogeneous
language communities.  The purposive selection spans scripts and documented
model-support conditions; it is not a probability sample of the world's
languages, dialects, or users.

All natural-language fields are stored in Unicode NFC. Protected identifiers,
JSON keys, schema delimiters, code literals, and canonical answers use a fixed
byte-stable ASCII machine schema;
surrounding prose, word order, punctuation, and spacing follow the registered
written variety.  Locale, script, normalization, digit, and punctuation rules
are frozen in the language manifest.  A post-freeze orthography change is a
protocol deviation rather than a silent correction.

Terms such as ``high resource'' and ``low resource'' are not primary treatment
labels.  Any exploratory resource index is frozen from documented corpus and
model-support variables without outcome access.  Because only seven
non-English languages are purposively selected and checkpoints do not expose
complete training mixtures, the index is descriptive and cannot support a
population-level resource or training-exposure claim.

\subsection{Rendering protocol}

A rendering will consume a typed item and produce a prompt without changing its
variables, quantifiers, negations, entity relationships, answer domain, or
difficulty label.  Numeric and symbolic slots remain machine identifiable.
The renderer emits a structured alignment record connecting every
natural-language clause to its latent predicate.  Programmatic checks confirm that:

\begin{itemize}
  \item every required slot occurs exactly once unless repetition is declared;
  \item no unregistered number, entity, edge, or ordering relation is added;
  \item polarity and quantifier annotations match the latent representation;
  \item the canonical answer remains representable in the fixed answer schema;
  \item Unicode normalization does not change a protected identifier.
\end{itemize}

The main corpus will not be constructed by translating every English sentence.
Where the typed representation permits direct realization, languages are
rendered independently from the same structure.  When a natural scenario
requires adaptation, its source path is assigned before model execution by a
frozen schedule balanced over target language, task family, and difficulty.
The source/adaptation path is recorded, included as a registered nuisance term,
and never selected using model outcomes.  Independent rendering from the typed
representation is designed to limit systematic artifacts introduced by translation
~\cite{artetxe2020translation}; outcome-blind scenario adaptation separately
addresses the English-centered cultural assumptions identified by Macaron
~\cite{elsetohy2026macaron}.

\subsection{Human review and adjudication}

Each unique confirmatory language-template version, the shared RQ3
language-codebook, and each localized RQ3 answer-template version will be reviewed
independently by two qualified reviewers. A qualified reviewer will have documented native or professionally
verified near-native literacy in the registered written variety under a frozen
rubric.  Reviewers will judge grammaticality, semantic equivalence, naturalness,
answerability, cultural assumptions, terminology, and whether the registered
difficulty is preserved.  Disagreements will be resolved by a third equally
qualified reviewer who does not see model outcomes.  The future results will
report agreement and adjudication rates by language and category.

Every item-specific rendering must be produced from an approved template and
pass the deterministic slot, predicate, polarity, answer-domain, and Unicode
checks above. Reviewers inspect a frozen boundary-case instantiation suite for
every slot type; item-specific instances are not silently substituted for the
registered template version. This makes the human review unit and the
all-instance programmatic verification unit explicit.

RQ3 answer-text templates are part of the same review unit. For each task
family and registered output language, an approved controlled short-answer
grammar binds the localized predicate and its protected answer slot. The
machine-stable \texttt{answer\_code} remains ASCII; the localized
\texttt{answer\_text} is accepted only when its template, language, extracted
slot, code, and independent solver agree. This prevents the output-language
treatment from degenerating into eight copies of an identical ASCII wrapper.

Before any template review or manual trace audit, an ethics-governance receipt
will record the applicable institutional determination, consent basis and
version, compensation and payment timing, data minimization, confidentiality
and access controls, withdrawal deadline and handling, retention/deletion
schedule, content warning, and public small-cell suppression rule.
Review identities, contact information, and compensation records will remain in
a private governance store.  Public evidence will contain template identifiers,
redacted review decisions, adjudicated status, and digests.

Every candidate template version and human edit receives a content digest and
versioned provenance record. A registered evaluation checkpoint may not
generate, rank, or select a confirmatory template or rendering, and only the
adjudicated version may enter the frozen corpus.

\subsection{Public validation strata}

The paired synthetic corpus is planned to support identification; public
benchmarks will assess bounded external validity. MMLU-ProX supplies expert-reviewed multilingual
knowledge questions~\cite{xuan2025mmluprox}, PolyMath supplies multilingual
mathematics with difficulty levels~\cite{wang2025polymath}, and Macaron supplies
human-written multilingual and multicultural reasoning scenarios
~\cite{elsetohy2026macaron}.  Each will be analyzed under its own license,
translation provenance, answer format, and contamination limitations.  Public
benchmark rows will never be pooled into the primary paired estimand.
\section{Models, Runtimes, and Measurement}
\label{sec:models-runtime}

\subsection{Model-lineage contract}

The confirmatory study will register exactly three open-weight checkpoints from
three distinct training lineages: (i) a multilingual checkpoint with documented
thinking and non-thinking modes, (ii) a separately trained or distilled
reasoning checkpoint, and (iii) an independently developed dense reasoning
checkpoint outside the first two lineages.  Primary checkpoints will be dense models
in a reproducible 7B--14B range.  Each must expose an assignable observable
trace and exact trace/answer token boundaries.  Larger models,
mixture-of-experts models, and checkpoints with only hidden or aggregate
reasoning fields will be confined to separately labeled robustness strata.

Candidate selection will be completed before confirmatory outcomes are
generated.  For every checkpoint, the registry will record repository and
revision, weight digest, architecture class, parameter count, license,
tokenizer files, special-token mapping, chat template, precision, quantization,
observable-trace semantics, and documented language support.  The Qwen3
technical report, DeepSeek-R1 article, and Microsoft technical report for
Phi-4-reasoning document candidate thinking behavior and training lineage
~\cite{qwen32025report,deepseek2025r1,abdin2025phi4reasoning}.  In particular,
the Microsoft report describes Phi-4-reasoning as a 14B reasoning model with
detailed generated reasoning chains.  Its documented thinking block and
answer-summary interface make it a candidate for the observable-span contract,
but only a runtime token-level delimiter test can establish eligibility.  Qwen3
and Phi-4-reasoning are cited here as unreviewed technical-report/preprint
sources for experimental-object and interface facts only, not as peer-reviewed
evidence or novelty claims.

The current pre-download candidates are Qwen/Qwen3-8B,
deepseek-ai/DeepSeek-R1-Distill-Llama-8B, and
microsoft/Phi-4-reasoning; exact source commits are held in the model registry.
Their local weight and runtime digests do not yet exist, so these candidates are
not frozen experimental objects.  The registry preserves the pre-download replacement of
Mistral-Nemo-Instruct-2407 by Phi-4-reasoning after the former failed the
outcome-blind observable-trace eligibility gate; no pilot or study outcome
informed that decision.

The locally available Qwen3 30B MoE quantization will remain restricted to
infrastructure preflight.  Its records are marked
\texttt{claimEligible=false} and cannot enter a result table.  Local
availability will not determine the scientific model set.

\subsection{Runtime and span contract}

Each runtime adapter will receive the same typed request and emit a common
record.  The record will include the full serialized prompt, exact prompt and
generated token-ID arrays, registered span index sets, observable trace and
answer boundaries, monotonic timestamps, terminal reason, memory measurements,
and raw-response digest.  Span counts will be lengths of disjoint ranges in the
complete runtime arrays, not sums from independently tokenized fragments.  The
adapter must enforce three independently assigned observable trace-token
budgets, a trace--answer delimiter counted outside those budgets, deterministic
forced handoff, and a fixed answer-token reserve identical across conditions.
A runtime that exposes only aggregate completion counts, hidden reasoning, or
no reliable trace--answer boundary cannot support the primary trace analysis;
it may appear only in a diagnostic or external-validity stratum with the field
reported as unobserved.

RQ1--RQ2 hardware measurements will use stateless requests, batch size one,
fixed precision, disabled cross-request prefix caching and speculative
decoding, fixed threads and kernels, and a committed interleaved schedule.
Sampling is disabled with temperature zero and top-\(p\) one; the registered
seed binds provenance rather than serving as an inferential replicate. Each
checkpoint must pass a byte-identical repeat qualification before freeze. The
model will remain loaded between requests.  Load time will be reported
separately and excluded from per-request inference.  Warm-up requests will
continue only until the registered anchor stability rule is satisfied.

RQ3 is the sole registered exception to the no-state-carryover rule. A shared
mapping from eight equal-status fixed-width ASCII codes to the registered
output languages must pass qualified two-reviewer validation before design
freeze; it is then serialized before the trace and becomes part of
the sealed parent. Stage A
must complete the observable trace, append and separately count the common
structural delimiter, then stop at the frozen control-injection boundary before
an output-language control or answer token, and seal the immutable
serialized prefix token IDs, runtime-native KV state, both digests, runtime
identity, and parent-state identifier. Stage B restores eight byte-identical
copy-on-write clones in randomized order. Each branch then append-prefills the
same fixed-form ASCII selector template with exactly one registered
language-code value; the adapter records its exact token IDs/span and timestamps. Every
branch uses the parent checkpoint's deterministic greedy profile, the same
fixed \(a_0\) answer reserve, and the same terminal precedence.
The adapter must prove that branch execution cannot mutate the parent,
siblings, sampler, or session state. Exact text replay in fresh requests is a
robustness path, not an eligible substitute for the primary estimand based on
identical pre-treatment states.

Every attempt will be append-only. The scientific boundary begins when the
fully serialized, treatment-bound request enters content-dependent runtime
admission. Context overflow, input/content rejection, timeout, malformed answer,
OOM, answer-reserve exhaustion, runtime error/crash, and operator-cancelled
deviation after that boundary remain failure-inclusive terminal outcomes;
assigned-language noncompliance is a separate process outcome retained by
intention-to-treat. Trace-budget exhaustion is likewise a process event before
forced handoff and may coexist with a correct final answer. Only a demonstrably
treatment-independent transport, host, or bootstrap failure before
content-dependent admission has no scientific outcome; its record remains, and
at most one registered replacement is added under a new attempt identity. No
attempt overwrites a predecessor. Evidence-integrity failure places the
affected claim set on deviation hold rather than being recoded as model error.

\subsection{Time and memory}

The runtime will use a monotonic clock for dispatch, content-dependent
admission, inference entry, first generated token, trace end, RQ3
control-prefill start/end, answer end, and terminal response. Primary model wall
time starts at content-dependent admission; pre-admission transport/bootstrap
overhead is descriptive. The study will report the
registered first-token composite, trace and answer decoding time where their
boundaries are reliable, and total wall time. Engine
version and every latency-affecting optimization will be frozen because
framework overhead can dominate nominal compute differences
~\cite{fernandez2023framework}.

Peak allocated memory and KV-cache bytes will be recorded only when the runtime
exposes them without profiler-induced synchronization.  Profiler-only
measurements will run in a separate diagnostic stratum and will not be pooled
with unprofiled latency.

\subsection{Operating-system-reported energy measurement}

The planned primary path on Apple hardware will invoke \texttt{powermetrics}
only through an explicit operator-approved privileged command.  It samples an
operating-system estimate over the rails named in the frozen qualification
receipt; it is neither total wall-plug power nor a calibrated physical energy
meter. Sampling will begin before dispatch and end after the terminal
event so that the registered integration interval is fully covered. H3 gross
energy is integrated from content-dependent admission to terminal; H4 incremental energy
is integrated from branch control injection to branch terminal. Each
measurement block will record at least sixty seconds of idle power
before and after the block, fixed anchor prompts, sample timestamps, lost-sample
counts, host state, and the registered numerical integration rule.

Same-host total-request gross OS-reported energy may enter H3 only if a
pre-freeze validation demonstrates adequate sample count, timestamp alignment,
repeatability, and coverage for the registered request durations.  RQ3
incremental branch energy requires a distinct H4 qualification gate.  Failure
of either gate removes that endpoint before freeze and forces regeneration of
the corresponding power and multiplicity artifacts.  Net energy after idle
subtraction is secondary.  The ordinary sampling period cannot be assumed to
resolve a short prefill phase: prefill/decode phase estimates are exploratory
unless a separate stricter phase-resolution gate passes.

All energy comparisons will be within the same fixed host, operating state,
runtime, and measurement boundary. Estimates from different devices, Apple
operating-system rail estimates, and discrete-GPU board power will never be pooled or described
as directly comparable.  The paper will report the boundary, sampling period,
integration, idle adjustment, hardware, operating system, driver/runtime, and
validation receipt, following systematic energy-reporting guidance
~\cite{henderson2020energy} and phase-alignment concerns documented by
TokenPowerBench~\cite{niu2026tokenpowerbench}. A block failing the anchor or
alignment gate follows the field-level missingness and diagnostic-only recovery
rule in \secref{sec:experimental-design}; model outcomes are not replaced.

\subsection{Trace-language measurement}

Language compliance will be estimated with an ensemble of script proportions,
language identification, code-switch segmentation, and stratified manual
review.  Mathematical notation, code, identifiers, and the canonical answer
will be excluded before classification.  The detector will be calibrated
separately for each registered written variety and may return unknown or mixed
rather than forcing a language.

Classifier error will be propagated through a sensitivity analysis but will
not change the \ITT{} assignment.  For Latin-script pairs such as English,
Spanish, and Swahili, script detection alone will never count as language
evidence.

\subsection{Evidence and privacy boundary}

Raw traces may contain benchmark text, model-license restrictions, or
unexpected sensitive strings.  They will remain in a private,
access-controlled evidence root.  The paper repository will store content
digests, structured measurements, scorer outputs, and sanitized excerpts
approved for release.  No API credential, reviewer identity, hardware serial
number, private path, or unapproved trace will be included in the manuscript
package.
\section{Experimental Design and Analysis}
\label{sec:experimental-design}

\subsection{Research questions and hierarchy}

\begin{description}[style=nextline]
  \item[RQ1: Input representation (primary).] At fixed checkpoint, assigned
  observable-trace policy, canonical output, and the registered middle trace
  budget \(b_M\),
  how will \(\Lin\) change pass@1, exact prompt-token spans, first-token
  attainment, and restricted TTFT?
  \item[RQ2: Trace assignment (primary).] At fixed input, canonical output, and
  each of the three assigned observable trace-token budgets,
  how will assigned \(\Ltrace\) change pass@1, unconditional observable trace
  tokens, total generated-sequence tokens (trace plus delimiter plus answer),
  content-admission-to-terminal wall time, and qualified same-host gross
  operating-system-reported energy estimates over prespecified hardware rails?
  \item[RQ3: Answer realization (primary).] From byte-identical clones of one
  sealed pre-treatment prefix and runtime-native KV state, how will eight fully
  crossed fixed-form \(\Lout\) selectors change answer tokens,
  decode/total branch time, qualified incremental energy, format validity, and
  semantic preservation?
  \item[RQ4: Registered-set heterogeneity (descriptive).] How will RQ1--RQ2
  estimates differ across the fixed budgets, difficulties, task families,
  languages, and checkpoints?
  \item[RQ5: Multi-attempt policy (randomized secondary).] At the same total
  assigned observable trace-token allowance, how will a stochastic
  one-long-attempt policy differ from a complete \(K\)-short-attempt policy under frozen
  seeds and an oracle-blind aggregation rule?
  \item[RQ6: Observable behavior (descriptive).] How often will checkpoints
  mix or switch languages, repeat, self-correct, truncate, exhaust the assigned
  trace budget, or terminate before a valid answer?
\end{description}

RQ1--RQ3 define the five primary omnibus tests H1--H5 in one Holm family. RQ5
is a separately frozen randomized secondary experiment with its own Holm
family. RQ4 and RQ6 are descriptive and cannot rescue a primary claim or
support population inference about resources, lineages, scales, or world
languages.

\FloatBarrier
\begin{table*}[t]
  \centering
  \caption{Research-question, estimand, endpoint, and test registry.}
  \label{tab:rq-analysis-map}
  \small
  \begin{tabularx}{\textwidth}{L{11mm} L{31mm} L{42mm} Y L{18mm}}
    \toprule
    RQ & Contrast & Endpoint & Estimator/test & Family \\
    \midrule
    RQ1 & C1--C3, C2--C4 at \(b_M\) & pass@1, exact prompt spans,
      first-token attainment, restricted TTFT & paired
      max-\(\lvert t\rvert\) omnibus over frozen components & H1 \\
    RQ2 & C1--C2, C3--C4 at \(b_L,b_M,b_H\) & pass@1 operating points &
      categorical-budget paired omnibus & H2 \\
    RQ2 & same contrasts and budgets & failure-inclusive observable trace and
      total generated-sequence tokens, wall time, qualified gross energy & paired cost-component
      max-\(\lvert t\rvert\) omnibus & H3 \\
    RQ3 & eight branches from an exact parent prefix/KV clone & answer tokens/
      decode, branch time, energy, format/preservation; selector cost descriptive & blocked
      fully crossed max-\(\lvert t\rvert\) omnibus & H4 \\
    RQ2 & trace-policy contrast across \(b_L\) versus \(b_H\) & pass@1
      difference-in-differences & paired interaction omnibus & H5 \\
    RQ5 & one long versus \(K\) shorter attempts at equal total trace allowance
      & frozen success and cost components & oracle-blind randomized analysis &
      separate secondary \\
    RQ4, RQ6 & fixed strata and observable behavior & absolute outcomes and
      intervals & descriptive registered sensitivities & none \\
    \bottomrule
  \end{tabularx}
\end{table*}

H1--H5 each yield exactly one prespecified resampling omnibus \(p\)-value.
These five values form one Holm family controlling family-wise error at 0.05.
One additional bootstrap max-\(\lvert t\rvert\) critical value over the union
of all preregistered H1--H5 components constructs a global 95\% simultaneous
component band. A named endpoint, language, checkpoint, or budget claim
requires both a Holm-rejected H-level omnibus and its interval in that global
band to exclude the applicable null or decision margin. No local alpha or
component subset is selected after outcomes. The frozen registry contains
every coefficient vector, finite-set weight, tail, smallest effect size of
interest (SESOI), margin, covariance estimator, and seed.

\subsection{Conditions, trace budgets, and run count}

The schedule covers exactly EN, ZH, ES, AR, HI, JA, SW, and TH. At each
item--checkpoint pair, C1--C3 run for seven non-English languages and C4 runs
once as the shared English/English baseline, giving \(7\times3+1=22\) unique
input- and trace-language conditions. C4 is referenced rather than duplicated, and
its shared covariance is retained in paired inference.
The pilot will freeze exactly three assigned observable trace-token budgets,
\(b_L,b_M,b_H\), and one separate fixed answer-token reserve \(a_0\). The
structural trace--answer delimiter is counted separately and consumes neither
allowance. Every item--checkpoint pair receives all three trace budgets, and
all 22 input- and trace-language conditions are run within every item--checkpoint--budget
block. A committed permutation maps each atomic
\((i,m,c,b,\mathrm{seed})\) identity to one thermal execution slot, balanced
within checkpoint, task-family, and registered-difficulty strata. Run order is
randomized within thermal blocks and is not a scientific treatment.

\begin{equation}
  240\times3\times3\times(7\times3+1)=47{,}520
  \label{eq:planned-run-count}
\end{equation}

is the planning single-seed RQ1--RQ2 schedule. Pilot runs, primary RQ3
continuations, repeated-seed robustness, diagnostic arms, and randomized RQ5
multi-attempt policy runs use separate namespaces and receipts and are outside
this count.

The core and RQ3 schedules use the same frozen greedy decoding profile with
sampling disabled, temperature zero, and top-\(p\) one. The single seed binds
serialization and execution provenance; it is not a sampled model factor.
Byte-identical repeat qualification is required for each checkpoint before
design freeze.

\begin{table*}[t]
  \centering
  \caption{Planned run-accounting layers. Pilot, qualification, and robustness
  runs use separate namespaces and are excluded.}
  \label{tab:run-accounting}
  \scriptsize
  \begin{tabularx}{\textwidth}{L{18mm} Y Y Y}
    \toprule
    Layer & Finite frame & Executed requests or continuations & Inferential role \\
    \midrule
    RQ1--RQ2 core & $N\times3$ checkpoints $\times3$ budgets $\times22$ cells
      & $198N$; $47{,}520$ at planning $N=240$ & H1, H2, H3, H5 \\
    RQ3 & $198n_3$ parent states
      & $12n_3$ sampled parents; up to $96n_3$ Stage-B branches & H4 with frozen inclusion weights \\
    RQ5 & $n_5\times3$ checkpoints $\times22$ input- and trace-language conditions
      & $66n_5(1+K)$ requests: one long and all $K$ short attempts per unit
      & separate S1--S2 family \\
    \bottomrule
  \end{tabularx}
\end{table*}

\subsection{Pilot and power}

A disjoint pilot will use an item set selected before any model outcome to
select the finite trace-budget grid, the middle budget, and the fixed
answer-token reserve; validate timeout,
warm-up, and thermal controls; estimate paired
discordance and dependence, choose qualified positive-cost models, and test
the power instrument. They may not select languages, checkpoints, cells,
hypotheses, directions, or reporting thresholds.

Power simulation will reproduce the complete three-budget crossing, 22-cell
schedule within every budget, shared-C4 covariance, item and thermal-block
dependence, Stage-B eligibility/blocking from the sealed pre-treatment state, H1--H5 Holm
procedure, and global component band. The core and RQ3 item counts will be the
smallest for which the Monte Carlo 95\% lower confidence bound reaches 90\%
both for every Holm-adjusted H-level omnibus under its least-favorable
registered smallest effect size of interest and for every frozen headline
decision event: H-level rejection plus the global-band interval excluding its
applicable null or margin. Headline targets are equal-weight finite-set
aggregate components. Other components must meet a frozen interval half-width
ceiling or be labeled precision-limited.
Initial planning values are a two-percentage-point pass@1 smallest effect of
interest, a 1.10 multiplicative positive-cost mean-ratio smallest effect of
interest, positive-cost equivalence bounds \([1/1.10,1.10]\), and a
\(-2\)-percentage-point accuracy non-inferiority margin. Simulation count,
convergence rules, missing-instrument/eligibility scenarios, energy and latency
fallback branches, margins, and any revision to 240 or \(n_3\) items freeze
before \texttt{DESIGN\_FROZEN}.

The executable precollection audit retains \(N=n_3=240\) only as an initial
corpus receipt and performs a bounded 12-item-increment sensitivity search.
Under its synthetic standardized planning assumptions, the first lexicographic
boundary is \(N=2352,n_3=420\), with a weakest Monte Carlo 95\% power lower
bound of 0.901 across the registered baseline, dependence, and instrumentation
branches. This is a design diagnostic, not a selected sample size: the
standardized alternatives do not instantiate the natural-scale ratio,
non-inferiority, or equivalence margins, and no disjoint pilot input is
available. The audit therefore records \texttt{selectedCounts=null}. Final
counts require independently replayed pilot-bound covariance, eligibility, and
per-component scale, direction, smallest-effect, and margin registries; failure
within the declared resource cap keeps collection on hold rather than relaxing
the scientific gate.

If total-request energy fails its coverage, alignment, cadence, repeatability,
or anchor gate, it is removed from H3 before freeze and the power and
multiplicity artifacts are regenerated. Net and phase energy remain secondary
or exploratory as specified in Section~\ref{sec:models-runtime}.

\subsection{Primary answer-language intervention from identical pre-treatment states}

RQ3 begins only after its own design, eligibility, power, and H4 multiplicity
artifacts freeze. A disjoint \(n_3\)-item target frame crosses all 22
input- and trace-language conditions, three trace budgets, and three checkpoints. For every
item--checkpoint pair, an outcome-blind restricted draw selects one of the 21
non-English-language--budget pairs and executes its corresponding C1--C4
parents. The frozen schedule balances pairs within
checkpoint--family--difficulty strata and across checkpoints. Specified
C1--C3 parents have inclusion probability \(1/21\), whereas C4 at a specified
budget has probability \(1/3\); Horvitz--Thompson/Hájek weights freeze before
generation. Thus the target frame has \(198n_3\) parents, the executed panel
\(12n_3\), and the assigned eight-branch continuation set \(96n_3\). An eligible
Stage-A run completes its trace, appends and separately counts the common
structural delimiter, then stops at the registered control-injection boundary
before any output-language control or answer token. The runner seals the
complete serialized prefix-token sequence, runtime-native immutable KV
snapshot, their digests, parent-state and checkpoint/runtime identities, eight
branch descriptors, and randomized branch order.

The sealed parent contains one shared mapping from eight equal-status
fixed-width ASCII codes to the registered output languages after that mapping
has passed qualified two-reviewer validation before design freeze. Every
output-language arm restores a byte-identical copy-on-write clone and then
appends the same fixed-form ASCII selector template with only the registered
code changed. Exact selector token IDs and spans are recorded. No arm may inherit sampler state,
conversation history, mutable cache changes, or other branch state. The
identical-state claim therefore concerns the randomized pre-treatment branch
point, not the post-control KV states.

Each branch emits the same frozen two-field schema: ASCII
\texttt{answer\_code} plus a task-family-specific localized
\texttt{answer\_text}. The preservation scorer requires code correctness,
localized-template validity, and surface-slot/code/solver agreement; a shared
ASCII payload alone cannot satisfy H4.

H4 targets the observed pre-treatment-eligible states in the frozen target
frame using frozen design weights. Sampled ineligible states remain in the
attrition ledger and are not replaced; reporting separates frame, sample,
runtime-entry, boundary, eligibility, branch, terminal, and energy-observed
denominators. Selector tokens and prefill time are reported as descriptive
design diagnostics and do not enter H4. H4 endpoints are answer tokens and
decode time, total branch wall time, qualified
incremental energy from control injection through terminal, format validity,
and solver-verified semantic preservation. Every branch receives the same fixed
\(a_0\) answer reserve and frozen terminal precedence. Fresh-request serialized-text replay
is a labeled robustness analysis: it repeats prefill and is never pooled with,
or used to rescue, the result based on identical pre-treatment states. Neither design estimates an
end-to-end effect of declaring output language before reasoning.

\subsection{Randomized secondary multi-attempt policy experiment}

RQ5 uses disjoint secondary items and crosses both complete policies for each
item, checkpoint, and input- and trace-language condition at one pilot-frozen total trace
allowance \(T_5\); execution order is randomized. The long policy receives
\(T_5\) in one attempt. Both policies use the frozen stochastic decoding
profile (temperature 0.6, top-\(p\) 0.95). Before any output, each unit receives
a frozen \(K\)-seed vector: the long policy uses \(s_1\), while the short policy
uses \(s_1,\ldots,s_K\) in its registered attempt order. The \(K\)-short policy
executes every attempt, with integer allowances that sum exactly to \(T_5\), and
each attempt receives its own fixed answer reserve. The oracle-blind aggregator applies the frozen parser,
selects the plurality of non-malformed normalized candidates, breaks ties by
precommitted randomized attempt rank, and emits the registered null output if
none is valid. It cannot access the gold answer, solver result, correctness
label, or any outcome-adaptive instruction.

Accordingly, RQ5 estimates a complete multi-attempt policy contrast at equal
total trace allowance. The short policy has up to \(K\) answer opportunities
and \(K a_0\) potential answer capacity, whereas the long policy has one of
each. All realized answer tokens and repeated prompt/prefill work enter S2, but
unused reserve is not a realized cost. No result will describe RQ5 as a pure
trace-allocation effect or as equal-total-compute evidence.

RQ5 latency and energy require a separate policy-level qualification gate; the
single-request H3 and branch-level H4 gates do not establish eligibility for a
multi-request policy. Before RQ5 output, the gate validates monotonic timestamp
alignment, sampling coverage/cadence, repeatability, sampler overhead, and the
aggregation segment for both the long and complete \(K\)-short policies.
Failure activates the prospectively frozen S2 component fallback.

S1 tests paired aggregate success. S2 tests the paired full-cost vector:
repeated prompt tokens/prefill, total observable trace tokens, total generated
tokens, wall time, and qualified energy, with the latter two including
aggregation computation through registered aggregate-output emission. S1 and
S2 each produce one max-\(\lvert t\rvert\) omnibus p-value and form a separate
two-value Holm family at 0.05. Named secondary components require one global
95\% max-\(\lvert t\rvert\) band across the S1--S2 vector. A separate power simulation
freezes \(K\), \(T_5\), secondary item count, allocation arithmetic, parser,
tie rule, margins, and missing-cost fallback. Policy-specific wall-time
\(W_{5,\max}\) and energy \(E_{5,\max}\) bounds include every request plus
aggregation. Missing S2
time or energy uses least-favorable policy-level bounds and the supremum
\(p\)-value over admissible completions in both S2 and its global band. Failure
of a bound, coverage gate, or independent certificate removes only that S2
component, leaves the two-value Holm family unchanged, and prohibits an
unqualified secondary efficiency claim. The secondary power simulation
reproduces every such branch. RQ5 cannot change H1--H5.

\subsection{Outcomes and failures}

Only a solver-verified canonical answer is pass@1 success. The scientific
boundary begins when a serialized treatment-bound request enters
content-dependent runtime admission. Every such request receives one terminal
class under a frozen precedence rule: context overflow, input rejection,
content-policy rejection, OOM, timeout, answer-reserve exhaustion without a
valid answer, completed malformed, completed incorrect, completed correct,
runtime error, runtime crash, or operator-cancelled deviation. Trace-budget exhaustion
is a separately recorded nonexclusive process event, not a competing terminal
class; it is not itself a failure when the frozen handoff rule yields a valid
answer within \(a_0\). Every non-correct terminal
class remains a failure in the intention-to-treat denominator. Partial tokens,
observed wall time, and qualified gross energy remain in unconditional cost
outcomes.

Trace and answer counts may be zero and use a two-part model. First-token
attainment and restricted TTFT form a two-part endpoint. H3 uses the composed
unconditional count mean \(\Pr(Y>0)\mathrm{E}[Y\mid Y>0]\) and its paired
natural-unit difference over all model outcomes; zero/positive parts are
mechanism summaries, not extra H3 components. Count ratios require both
unconditional means to be strictly positive. Positive wall time
and qualified gross energy use a pilot-frozen Gamma or log-normal mean model
plus natural-unit paired differences. Corrupt instrumentation makes only that
field missing; valid task and token fields remain, energy is never imputed
from tokens, and negative idle-adjusted net energy is retained in its
secondary natural-scale analysis. For claim-eligible energy, a pre-outcome
finite upper bound \(E_{\max}\) is frozen from the qualified platform-power
ceiling and request timeout. Missing energy is propagated through worst-case
\([0,E_{\max}]\) partial-identification bounds and through the global component
band. When allowed missingness remains, the energy component and H-level
omnibus use the supremum \(p\)-value over all admissible completions, with an
independent optimization certificate. A named energy or efficiency claim
requires every extreme-case bound to qualify. If \(E_{\max}\), the frozen
coverage gate, or the certificate fails, a prespecified fallback removes
energy from H3/H4 and efficiency claims, retains one H3 omnibus over its frozen
non-energy components, and reports complete-instrument energy only
descriptively with cell-wise missingness.

The same rule covers corrupted latency fields with the frozen timeout as
\(T_{\max}\): least-favorable \([0,T_{\max}]\) bounds and a supremum omnibus
\(p\)-value apply when the coverage gate allows missingness. Failure of the
bound, gate, or independent certificate removes only the affected H1/H3/H4
time components under the frozen fallback; it never deletes correctness,
token, or terminal outcomes and does not change the five-test Holm family. An
energy or latency fallback prohibits an unqualified efficiency claim.

All confirmatory outcomes are indexed by assigned trace budget. For every
registered realized cost \(C\), the failure-inclusive joint success--cost
estimand is

\begin{equation}
  G^{C}_{p,b}(c)=\Pr\!\left(A=1\ \land\ C\leq c
    \mid Z_{\mathrm{language}}=p,\ B=b\right).
  \label{eq:design-joint-success-cost}
\end{equation}

It is never replaced by \(\Pr(A=1\mid C\leq c,p,b)\), which conditions on a
post-treatment realized cost. Correctness-versus-realized-token, time, or
energy plots and empirical Pareto sets are descriptive and failure-inclusive;
they are not fixed-time or fixed-energy causal frontiers. An unqualified
efficiency statement requires a simultaneous accuracy lower bound above its
non-inferiority margin, at least one declared cost-ratio upper bound below
1.00, and every other declared primary-cost upper bound within its frozen harm
margin. Otherwise wording is endpoint-specific.

\subsection{Statistical models and dependence}

The primary estimator is the design-weighted paired finite-set contrast, with
the shared C4 retained once and semantic-item-by-thermal-block multiway
resampling preserving every within-item cell--budget identity. The exact
studentization, multiway resampling algorithm, number of draws, finite-set
weights, and covariance handling freeze in executable code before outcomes.
The following mixed model is a registered sensitivity analysis rather than the
source of primary claims:

\begin{align}
\operatorname{logit}\Pr(A_{imcb}=1)={}&\beta_0+\beta_{\mathrm{cell}}[c]
  +\beta_B[b]+\beta_m+\beta_{\mathrm{task}}+\beta_{\mathrm{difficulty}}
  +\beta_{\mathrm{render}} \nonumber\\
 &+(\beta_{\mathrm{cell}\times B})[c,b]
  +(\beta_{\mathrm{cell}\times m})[c,m]
  +u_i+v_{\mathrm{template}}.
  \label{eq:glmm}
\end{align}

Assigned trace budget is categorical. Checkpoints and languages are equally
weighted fixed registered factors; resource level and model scale are not confirmatory
moderators. Mixed models follow the procedures documented by Bates et
al.~\cite{bates2015lme4}. Latency/energy sensitivity models additionally include
thermal-block effects. Uncertainty uses semantic-item-by-thermal-block multiway
resampling, preserving all 66 cell--budget identities and the shared C4 within
each budget. Template/rendering terms and leave-one-task, language, and
checkpoint analyses are registered sensitivities.

\subsection{Exclusions, retries, and stopping}

Whole-record exclusion is limited to duplicate identity or a demonstrably
treatment-independent transport/host/bootstrap failure before content-dependent
admission. Content-dependent context/length/policy rejection remains a
scientific failure and is not replaced. Evidence-integrity or digest failure
places the affected claim set on deviation hold rather than being recoded as a
model outcome. Outcome-blind field corruption makes
only the affected latency or energy field missing; correctness, token, process,
and terminal evidence remain in their registered denominators. Noncompliance,
zero trace, malformed output, OOM, timeout, trace-budget exhaustion,
answer-reserve exhaustion, runtime error/crash, operator cancellation, and
truncation remain outcomes. One pre-admission replacement is allowed only for
the treatment-independent class under a new identity; both attempts remain
append-only. A scientific outcome is never rerun for improvement.

An anchor or alignment failure marks the affected block-level latency/energy
fields missing without deleting or replacing model outcomes. One recovery run
may be made under a new diagnostic-only identity to test instrument
restoration; it never enters the primary denominator or replaces the original
block. A second failure triggers the frozen infrastructure stop. Collection
never stops or expands for an interim effect or \(p\)-value. Analysis starts only after the
independent checker closes the complete three-budget crossing, all 66
cell--budget identities, the shared C4 within each budget, attempts, blocks,
instrument coverage, and every terminal state.
\section{Planned Results and Reporting Structure}
\label{sec:results}

\noindent\textbf{Evidence status: \StudyStage; empirical reporting
prospectively locked.}
This section predefines the reporting structure. Numerical tables and figures
activate only after the independent checker accepts the complete frozen
ledgers; empirical entries are generated exclusively from verified records.

The infrastructure-only v4 preflight verified the operational paths. All 40
scheduled records and 80 invocation-journal events were accounted for: 33
requests completed, seven reached their preflight generation limit, 31
responses were syntactically parsed, and none ended in infrastructure error.
The receipt states \texttt{operationalPreflight=true},
\texttt{claimEligible=false}, and
\texttt{tokenDecompositionValidated=false}. These permanently claim-ineligible
counts document operation of the request, append-only ledger, recovery,
receipt, and verification paths. Accuracy, language effects, span
decomposition, and model comparisons will be populated only from independently
verified confirmatory evidence.

\subsection{Collection integrity and treatment balance}

The first confirmatory table will account for every atomic identity by item,
checkpoint, input- and trace-language condition, assigned trace budget, decoding seed,
thermal block, attempt, process event, and terminal class.  It will verify all
three budgets and 22 cells, reuse shared C4 exactly once per
item--checkpoint--budget--seed block, and distinguish trace-budget exhaustion
from answer-reserve exhaustion without a valid answer.  Model, tokenizer,
template, runtime, scorer, language-review, energy, and design-freeze digests
will precede every task outcome.

The planned RQ1--RQ2 denominator is 47{,}520 scheduled invocations before any
prospectively power-frozen item-count revision.  Pilot, repeated-seed,
diagnostic, RQ3 branch, RQ5, and external-validation records will appear in
separate namespaces and denominators.

\subsection{H1/RQ1: input representation}

For each checkpoint and non-English language, the report will present C1--C3
and C2--C4 at the registered middle trace budget.  Absolute pass@1, exact
serialized prompt-token spans, first-token attainment, and restricted TTFT
will precede paired differences and ratios.  H1 contributes one omnibus
\(p\)-value to the primary Holm family; a named component requires its
registered simultaneous interval.  Tokenizer fertility remains an
associational mechanism measure rather than a separately randomized cause.

\subsection{H2, H3, and H5/RQ2: assigned trace policy}

The principal operating-point figure will compare C1--C2 and C3--C4 at
\(b_L\), \(b_M\), and \(b_H\). H2 covers pass@1. H3 covers
failure-inclusive unconditional observable trace tokens, total
generated-sequence tokens (trace plus delimiter plus answer),
content-admission-to-terminal wall time, and qualified same-host gross
OS-reported energy. H5 covers the
low-versus-high trace-budget difference-in-differences for assigned-trace
correctness; \(b_M\) is not part of that interaction.

All outcome classes remain in the intention-to-treat denominator.
Assigned-language compliance accompanies the treatment contrast, but compliant-only or
detected-language estimates are secondary.  Qualified energy is omitted, not
imputed, if its pre-freeze gate fails.  Each of H2, H3, and H5 contributes one
omnibus value to the same primary family.

For every registered cost \(C\), the report will display

\begin{equation}
  G^C_{p,b}(c)=\Pr(A=1\ \land\ C\leq c
    \mid Z_{\mathrm{language}}=p,B=b),
  \label{eq:results-joint-success-cost}
\end{equation}

alongside \(g_p(b)=\Pr(A=1\mid Z_{\mathrm{language}}=p,B=b)\).  It will not condition
correctness on observed \(C\leq c\).  Realized-cost Pareto sets and AUCs are
descriptive and receive no fixed-time or fixed-energy causal interpretation.
The threshold grid is frozen from pilot-only records; no inferential
curve-difference claim is registered. If energy is missing, \(G^E\) appears as
the preregistered lower/upper partial-identification band rather than a point
curve.

\subsection{H4/RQ3: answer-language intervention from identical pre-treatment states}

Before comparing output languages, the report will show the disjoint Stage-A
target frame, balanced-incomplete four-parent panel, frozen inclusion
probabilities/weights, and complete eligibility/attrition by cell, budget, and
checkpoint. It will reconcile \(198n_3\) target parents, \(12n_3\) sampled
parents, and at most \(96n_3\) assigned branches. Every entered state must stop at the control-injection
boundary and seal its serialized prefix token IDs, runtime-native KV snapshot,
prefix/KV digests, runtime identity, parent ID, and eight branch descriptors.
The checker must verify byte-identical pre-treatment copy-on-write restoration,
randomized branch order, the shared codebook and exact fixed-form selector token
span, and absence
of mutable cache, session, sampler, parent, or sibling leakage. Planned
ineligible states remain in the attrition ledger and are not replaced.

The report will separately show schema validity, ASCII
\texttt{answer\_code} correctness, localized-surface validity, and agreement
among the surface-extracted slot, code, and solver. Semantic preservation
requires all registered semantic checks; code-only success cannot pass it.

Selector tokens and prefill time will be shown as descriptive design
diagnostics. H4 will compare answer tokens/decode time, total branch wall time,
qualified incremental energy, format validity, and
solver-verified semantic preservation across the eight output-language branches
using one blocked omnibus value and registered simultaneous component intervals.
Fresh-request exact serialized-text replay
will appear only in a separately labeled robustness table; it is neither
pooled with H4 nor used to replace an ineligible branch from the sealed
pre-treatment state.

\subsection{Primary multiplicity and efficiency decisions}

H1--H5 will contribute exactly five omnibus \(p\)-values to one Holm family at
family-wise error 0.05.  The result table will show unadjusted values,
Holm-adjusted decisions, natural-unit effects, and intervals from one global
95\% max-\(\lvert t\rvert\) simultaneous band over the complete frozen H1--H5
component vector. Rejecting an omnibus is necessary but insufficient for a
named language, checkpoint, budget, or endpoint statement; the matching global
band interval must also qualify.

An unqualified efficiency statement additionally requires its simultaneous
accuracy lower bound to exceed the registered \(-2\)-percentage-point
non-inferiority margin, at least one declared positive-cost ratio upper bound
below 1.00, and every other primary-cost upper bound within its frozen harm
margin.  Otherwise the manuscript will report only qualifying
endpoint-specific effects.

\subsection{RQ4 and RQ6: descriptive registered-set results}

RQ4 will display absolute outcomes and treatment contrasts across the fixed
checkpoints, languages, task families, difficulties, rendering paths, and
trace budgets.  Equal finite-set weights, shared-C4 covariance, and registered
leave-one-group-out sensitivities will be explicit.  No result will be phrased
as a population effect of resource level, lineage, scale, or language
community.

RQ6 will report assigned-language compliance, mixing/switching, repetition,
self-correction, empty traces, trace-budget exhaustion, answer-reserve failure,
timeout, malformed completion, OOM, runtime error/crash, and incorrect completion
with absolute denominators.  Visible traces will not be described as a
complete account of internal computation.

\subsection{RQ5 and external validation}

The randomized secondary RQ5 table will compare a stochastic one-long-attempt
policy with a frozen complete \(K\)-short-attempt policy under equal total
assigned trace-token allowance, a registered \(K\)-seed vector, and an
oracle-blind aggregator. The long policy uses \(s_1\), while the short policy
uses \(s_1,\ldots,s_K\). The frozen rule may inspect candidate payloads but
not gold/reference answers, solver outputs, or correctness labels. Repeated
prompt/prefill work and every realized trace, delimiter, and answer token will
be counted as costs; unused answer-reserve capacity will not. Wall time and
qualified energy include aggregation computation through registered
aggregate-output emission. RQ5 uses its own
secondary Holm family and cannot alter H1--H5.
Because each short attempt has its own \(a_0\), the table will explicitly label
this a full-policy contrast with unequal potential answer capacity, not a pure
trace-allocation or equal-total-compute comparison.

Public benchmarks, hosted models, additional checkpoints, and repeated-seed
analyses will be reported separately with their own denominators, revisions,
licenses, and evidence labels.  They may assess external validity but cannot
rescue or overturn a primary decision.

This section becomes result-bearing only when every generated table, figure,
statistical object, and claim--evidence row identifies the same frozen manifest,
ledger, analysis configuration, and independent-checker receipt digests.
\section{Robustness and Interpretation}
\label{sec:robustness}

\subsection{Registered robustness analyses}

The primary analysis will be accompanied by the following prespecified checks:

\begin{itemize}
  \item leave-one-checkpoint-out and leave-one-task-family-out estimates;
  \item alternative log-normal and Gamma cost models selected without access
  to treatment coefficients;
  \item two-part zero/positive cost models and paired natural-scale net-energy
  intervals;
  \item semantic-item cluster bootstrap intervals and template-level
  sensitivity;
  \item multiway semantic-item--thermal-block resampling for latency and energy;
  \item qualified total-request gross OS-reported energy versus secondary
  idle-adjusted net energy, with failed instrument fields marked missing under
  the registered outcome-blind anchor rule and model outcomes retained;
  \item deterministic decoding versus a stratified repeated-seed subset;
  \item RQ3 estimates from identical pre-treatment states versus fresh-request
  exact serialized-text replay, with replay labeled robustness-only and never
  pooled with H4;
  \item direct-answer and free-choice trace policies;
  \item language-compliance measurement under alternative calibrated
  classifiers and manual-audit corrections;
  \item public-benchmark external validation analyzed separately from the typed
  synthetic corpus;
  \item dense primary checkpoints versus a separately labeled MoE robustness
  stratum; and
  \item natural-language output versus the canonical-answer primary stratum.
\end{itemize}

Robustness analyses will not promote a primary omnibus test that fails its Holm
gate or an efficiency comparison that fails its registered non-inferiority and
cost criteria. Leave-one-checkpoint and leave-one-task analyses will diagnose
dependence of the finite-set average on a registered group; they will not
silently replace its estimand or act as post-outcome subgroup-selection gates.
A directional aggregate will be called stable only when every registered
leave-one-checkpoint-out and leave-one-task-family-out estimate retains the same
direction and no simultaneous interval supports a material opposite effect. If
the sensitivity estimates cross zero but remain within the registered SESOI,
the claim will be downgraded to an average effect with registered
heterogeneity. A reversal beyond the SESOI will withdraw the aggregate
directional claim. The originally registered finite-set estimate and interval
will remain visible, and a favorable post hoc subgroup cannot rescue the
headline.

\subsection{How to interpret a language effect}

A measured input-language effect will be a property of a registered checkpoint,
tokenizer, task distribution, prompt template, and hardware/runtime setting.
It can combine representational length, lexical familiarity, training
exposure, and instruction following.  The paired design removes semantic-item
variation, but it does not randomize model pretraining.

An assigned-trace-language effect will be the deployable effect of a policy
instruction.  It includes whether the model obeys, mixes, changes length,
changes strategy, or fails to return a valid answer.  That policy effect is
operationally meaningful even when compliance is imperfect.  It is not the
causal effect of a hypothetical intervention that rewrites an otherwise
identical internal computation into another language.

Likewise, a tokenizer ratio will not be treated as a hardware-independent cost multiplier.
Input tokens principally affect prefill and memory, while generated tokens add
serial decode work; kernel and framework overhead can change the mapping from
tokens to time~\cite{fernandez2023framework}.  Energy depends on utilization,
precision, runtime, and workload composition
~\cite{fernandez2025energy,poddar2025sustainable}.  The paper therefore leads
with jointly measured outcomes rather than converting tokens to estimated
joules through a universal coefficient.

\subsection{Deployment implications}

Once the registered evidence gates are satisfied, the resulting estimates
support policy choices at three boundaries. A service can decide
whether to retain a user's input language, whether to assign or permit a pivot
language for generated reasoning, and how to realize the final answer.  The
right decision can vary by checkpoint, task difficulty, latency objective, and
fairness constraint.  A single globally cheapest language is neither assumed
nor sought.

Where a pivot policy improves an assigned trace-token-budget operating point,
deployment should still
report worst-language reliability and translation or compliance failures.  A
policy that saves compute by degrading a particular language outside the
non-inferiority margin is not described as an efficiency improvement.  This
distinction connects system efficiency to equitable access rather than treating
average cost as the only deployment objective.

\subsection{Relationship to tokenizer interventions}

The study will evaluate fixed checkpoint tokenizers; it does not propose a new
tokenizer.  Results may identify where parity-aware vocabulary construction
~\cite{foroutan2026parity} or cross-boundary segmentation
~\cite{tanase2026supratok} deserves evaluation, but they
cannot establish that a tokenizer retraining intervention would preserve the
same model behavior.  Such an intervention changes the vocabulary,
embeddings, training distribution, and sequence length jointly and requires a
separate controlled training study.
\section{Validity, Ethics, and Claim Boundary}
\label{sec:validity}

\subsection{Construct validity}

Token count is tokenizer-specific and does not equal FLOPs, time, energy,
price, or linguistic complexity.  The protocol will measure these quantities
separately.
Visible generated traces are not necessarily faithful explanations of a
model's prediction~\cite{turpin2023unfaithful}, nor do surface-language traces
establish the language of latent reasoning~\cite{liu2026latent}. In the title,
``reasoning efficiency'' is operationally defined as solver-verified task
success jointly with measured inference resources on reasoning tasks; it does
not denote access to private cognition. The body uses ``generated-trace
efficiency'' or ``inference efficiency'' for the measured mechanisms.

Language identification is difficult for short, mathematical, or mixed text.
The registered detector will exclude formulas and identifiers, permit unknown
and mixed labels, and be calibrated with a stratified human audit.  Assigned
language will remain the primary policy even when detection is uncertain.

\subsection{Internal validity}

Translation error could masquerade as a language effect.  Typed latent items,
clause alignment, invariant checks, two independent qualified reviews, and a
balanced, frozen source/adaptation-path schedule reduce this risk.  They cannot
make every rendering pragmatically identical, so rendering path and template
are modeled and leave-template-group-out checks are reported.  The locale,
script, Unicode, digit, and punctuation manifest prevents a broad language
label from silently hiding multiple written varieties.

Template-source choices can introduce source-specific phrasing or undeclared
dependence on an evaluated checkpoint. Complete content-digest, version, and
human-edit provenance is therefore required; a registered evaluation
checkpoint cannot generate, rank, or select a confirmatory template or
rendering, and no source bypasses the human review gate.

Thermal drift, cache state, background load, and runtime optimization can
confound time and energy. The experiment will randomize run order, interleave
languages, fix cache and batching, and record anchor prompts. A failed block
makes only its affected instrument fields missing; its correctness, tokens, and
terminal evidence remain. One diagnostic-only recovery block may test restored
instrumentation under a new identity, but it cannot replace primary outcomes; a
second failure activates the infrastructure stop. The
same hardware does not make checkpoint comparisons causal; primary language
effects are estimated within checkpoint.  Thermal-block terms and multiway
item-by-thermal-block resampling account for residual block dependence.

RQ3 additionally depends on exact pre-treatment-state identity. Each eligible
Stage-A prefix and runtime-native KV snapshot will receive independent digests,
and every one of its eight output-language branches will restore a
byte-identical copy-on-write clone in randomized order before append-prefilling its
registered fixed-form ASCII selector. A shared language-codebook must pass
qualified two-reviewer validation before design freeze and is then made part
of the sealed parent. The selector's exact token span, prefill
time, and incremental measurement boundary are retained. A branch that mutates
its parent or shares sampler/session state is ineligible. Serialized-text replay
in a fresh request repeats prefill and therefore remains robustness-only; it
cannot be pooled with or used to rescue H4.

Pilot-driven choices create researcher degrees of freedom. Pilot and
confirmatory seeds and roots will be disjoint. The three assigned observable
trace-token budgets, fixed answer-token reserve, delimiter accounting, atomic
thermal-slot schedule, sample size, exclusions, contrasts, SESOIs, H1--H5
omnibus tests, five-value primary Holm family, and analysis code will be frozen
before confirmatory generation. Pilot outcomes will never appear as evidence
in the abstract or conclusion.

\subsection{External validity}

The eight-language set covers several scripts and documented support conditions
but is not a representative sample of all languages, dialects, or users.  The
typed tasks
emphasize exact scoring and do not cover every creative, conversational, or
social use.  Public benchmark strata broaden task realism but reintroduce
translation, cultural, and contamination uncertainty.  Conclusions will
therefore remain conditional on the registered languages, checkpoints, tasks,
assigned observable trace-token budgets, fixed answer reserve, and hardware.

Hosted systems can update without notice and may conceal tokenization or
reasoning details. Any hosted-model validation records date, advertised
version, usage fields, and response digests; it supports external validity but
not the primary mechanism analysis.

\subsection{Statistical conclusion validity}

Translated versions of one semantic item are correlated. The primary finite-set
paired estimator and item resampling will
preserve all 66 cell--budget identities at each item--checkpoint pair and one
shared C4 within each of the three budgets. Mixed effects will model template
and item dependence; latency and energy will additionally include thermal-block
terms and multiway item-by-block resampling. One Holm procedure will protect
exactly the five prespecified H1--H5 omnibus tests
~\cite{holm1979sequential}, while randomized RQ5 attempt allocation will use a
separate secondary Holm family. RQ4 and RQ6 remain descriptive, and exploratory
comparisons use false-discovery-rate control
~\cite{benjamini1995fdr}. One global 95\% max-\(\lvert t\rvert\) band over the
complete frozen H1--H5 component vector prevents an omnibus rejection from
being rewritten as support for every language or endpoint. Equivalence
claims require a registered equivalence test~\cite{lakens2017equivalence}.  The
paper will report effect sizes, absolute denominators, and uncertainty rather than
relying on dichotomous significance language.

\subsection{Language equity and participants}

Language labels can invite essentialist interpretations.  We will not rank
languages or their speakers by cognitive efficiency.  A difference describes
a model and tokenizer operating on a registered task distribution, not an
intrinsic property of a language community.  Worst-language reliability and
between-language disparity will accompany aggregate efficiency.

Language reviewers are human participants in a limited annotation and
quality-control role. Before any template review or trace audit begins, a
hashed ethics-governance receipt must record the applicable institutional
determination, consent basis and version, compensation and payment timing,
data minimization, confidentiality and access controls, withdrawal deadline
and handling, retention/deletion schedule, content warning, and public
small-cell suppression rule. Reviewers must satisfy the frozen native or
professionally verified near-native literacy rubric for the registered written
variety. A timely withdrawal removes that person's not-yet-aggregated review
or audit data and invalidates affected receipts; replacement review follows
the same blinded qualification gate. Post-deadline handling follows the frozen,
disclosed consent and institutional rule and may invoke the preregistered safety
stop. Reviewer identity is never included in the public artifact.

\subsection{Environmental and reproducibility ethics}

The power study itself consumes energy.  Power-frozen item count, the
three-checkpoint primary tier, complete shared-reference accounting, and staged pilot reduce
unnecessary computation.  Failed and null results remain publishable under the
registered design, avoiding wasteful repetition for a preferred finding.
Energy reporting follows explicit measurement boundaries rather than
unsupported carbon conversions~\cite{henderson2020energy}. A qualified Apple
operating-system power stream will support, at most, a same-host estimate over
the rails named in the frozen receipt: it is neither total wall-plug energy, a
calibrated physical meter reading, nor
a cross-device comparison.  Idle-adjusted net energy will remain secondary,
and short-phase energy will remain exploratory unless its stricter resolution
gate passes.

The release package will include generators, redacted reviewed-template decisions,
locale and rendering-path manifests, sanitized machine-candidate provenance,
configuration manifests, scorers, independent verification, analysis code,
and derived records as licenses permit.  It will exclude credentials, reviewer
identities and compensation, raw hardware identifiers, private paths,
restricted benchmark text, and unapproved model traces. The publication gate
  will require named-author final-byte approval of all venue-required
  declarations, including funding and competing interests, author contributions,
  affiliation authorization, data/code availability, anonymization mode, and
  the release-scope review. Those declarations are finalized through the
  publication gate and remain separate from the protocol's scientific claims.

\subsection{Prohibited claims}

The evidence cannot support that one natural language is universally best for
reasoning, that surface trace language causes all internal computation, that a
token difference is a universal energy multiplier, that OS-reported energy is
total physical energy or cross-device comparable, or that the
registered sample represents all languages.  Such statements remain outside
the claim boundary even if a particular contrast is statistically significant.
\section{Conclusion}
\label{sec:conclusion}

Multilingual inference efficiency cannot be reduced to counting tokens in a
translated prompt. Input representation, assigned observable-trace language,
answer realization, task success, runtime latency, and same-host sampled
OS-reported energy estimates over prespecified hardware rails are distinct quantities. This
protocol is designed to separate them
with paired language-independent semantic items, deterministic answer oracles,
22 unique input- and trace-language conditions containing one shared C4 reference, and exactly
three assigned observable trace-token budgets \(b_L,b_M,b_H\) crossed within
every item--checkpoint pair. A separate fixed answer-token reserve \(a_0\) is
shared across conditions, and the structural delimiter is counted separately.
The initial planning single-seed RQ1--RQ2 schedule contains
\(240\times3\times3\times22=47{,}520\) runs across exactly eight registered
written language variants and three independently developed reasoning
checkpoints.

The primary analysis treats assigned trace language as a deployable policy and
retains noncompliance, truncation, content-dependent admission rejection,
malformed output, and post-admission failures
under intention-to-treat. It compares failure-inclusive outcomes at assigned
trace-budget operating points. Trace-budget exhaustion is not itself a failure
when the frozen handoff yields a valid answer within \(a_0\); answer-reserve
exhaustion without a valid answer is a failure. Wall time and qualified
same-host gross OS-reported energy remain measured outcomes, not randomized
fixed-time or fixed-energy budgets. Failure-inclusive joint success--cost
curves use \(G^{C}_{p,b}(c)=\Pr(A=1\land C\leq c\mid p,b)\), never correctness
conditioned on realized cost. H1--H5 each contribute one prespecified omnibus
\(p\)-value to a five-value Holm family, while named claims additionally require
their simultaneous intervals and decision margins.

Primary RQ3 will sample a frozen balanced-incomplete C1--C4 parent panel, then
execute eight fully crossed fixed-form ASCII language selectors in randomized
order after byte-identical copy-on-write cloning of each eligible sealed pre-treatment prefix
and runtime-native KV state, counting the control append-prefill cost and
requiring code/surface/solver agreement in a localized short-answer schema;
the schema and shared codebook must pass qualified review before freeze and are
then made part of the sealed parent, while fresh-request text replay remains
robustness-only. Randomized secondary RQ5
will compare a stochastic one-long-attempt policy with a complete
\(K\)-short-attempt policy at equal total trace allowance under frozen seeds, oracle-blind
aggregation, and a separate Holm family. Because potential answer capacity is
unequal, it is a full-policy comparison rather than a pure trace-allocation
effect. RQ4 and
RQ6 remain descriptive. Typed generation, two-reviewer language validation,
provenance records, immutable run ledgers, exact token-span accounting, and an
independently structured checker connect every eventual empirical statement to
auditable evidence. Final inferences remain scoped to the registered languages,
checkpoints, tasks, assigned trace budgets, runtime, and measurement host.

\StudyConclusionStatusSentence
 
\bibliographystyle{ACM-Reference-Format}
\setlength{\bibsep}{-0.2pt}
\bibliography{references}

\end{document}